\documentclass[fleqn,10pt]{wlscirep}
\usepackage[utf8]{inputenc}
\usepackage[T1]{fontenc}
\usepackage{makecell}
\usepackage{subcaption} 
\usepackage{pifont}
\usepackage{multirow}

\usepackage[natbib=true]{biblatex}
\usepackage{hyperref}

\title{Deep Age-Invariant Fingerprint Segmentation System}

\author[1, *]{M.G. Sarwar Murshed}
\author[1]{Keivan Bahmani}
\author[1]{Stephanie Schuckers}
\author[1]{Faraz Hussain}
\affil[1]{Department of Electrical and Computer Engineering, Clarkson University, Potsdam, NY 13699, USA}

\affil[*]{murshem@clarkson.edu}


\begin{abstract}
Fingerprint is one of the important modalities that have been used for biometric recognition applications such as border crossings, health benefits, criminal justice, electronic voting, etc. Fingerprint-based identification systems achieve higher accuracy when a slap containing multiple fingerprints of a subject is used instead of a single fingerprint. 
However, segmenting or auto-localizing all fingerprints in a slap image is a challenging task due to the different orientations of fingerprints, noisy backgrounds, and the smaller size of fingertip components.
The presence of slap images in a real-world dataset where one or more fingerprints are rotated makes it challenging for a biometric recognition system to localize and label the fingerprints automatically. Improper fingerprint localization and finger labeling errors lead to poor matching performance.  In this paper, we introduce a method to generate arbitrary angled bounding boxes using a deep learning-based algorithm that precisely localizes and labels fingerprints from both axis-aligned and over-rotated slap images.
We built a fingerprint segmentation model named CRFSEG (Clarkson Rotated Fingerprint segmentation Model) by updating the previously proposed CFSEG  model which was based on traditional Faster R-CNN architecture \cite{Murshed2021DeepSlapSeg}. CRFSEG improves upon the Faster R-CNN algorithm with arbitrarily angled bounding boxes that allow the CRFSEG to perform better in challenging slap images. 
After training the CRFSEG algorithm on a new dataset containing slap images collected from both adult and children subjects, our results suggest that the CRFSEG model was invariant across different age groups and can handle over-rotated slap images successfully. In the \textit{Combined} dataset containing both normal and rotated images of adult and children subjects, we achieved a matching accuracy of 97.17\%, which outperformed state-of-the-art VeriFinger (94.25\%) and NFSEG segmentation systems (80.58\%). The results indicate that the deep learning-based slap segmentation system is more efficient for both children and adult slaps.
Our pre-trained CRFSEG models and training codes for using our model are publicly available (\url{https://github.com/sarwarmurshed/CRFSEG}).
\end{abstract}

\begin{document}

\flushbottom
\maketitle
%
%
\thispagestyle{empty}


\section*{Introduction}
\begin{figure}
\begin{subfigure}[b]{0.40\textwidth}
\centering
\includegraphics[width=1\linewidth]{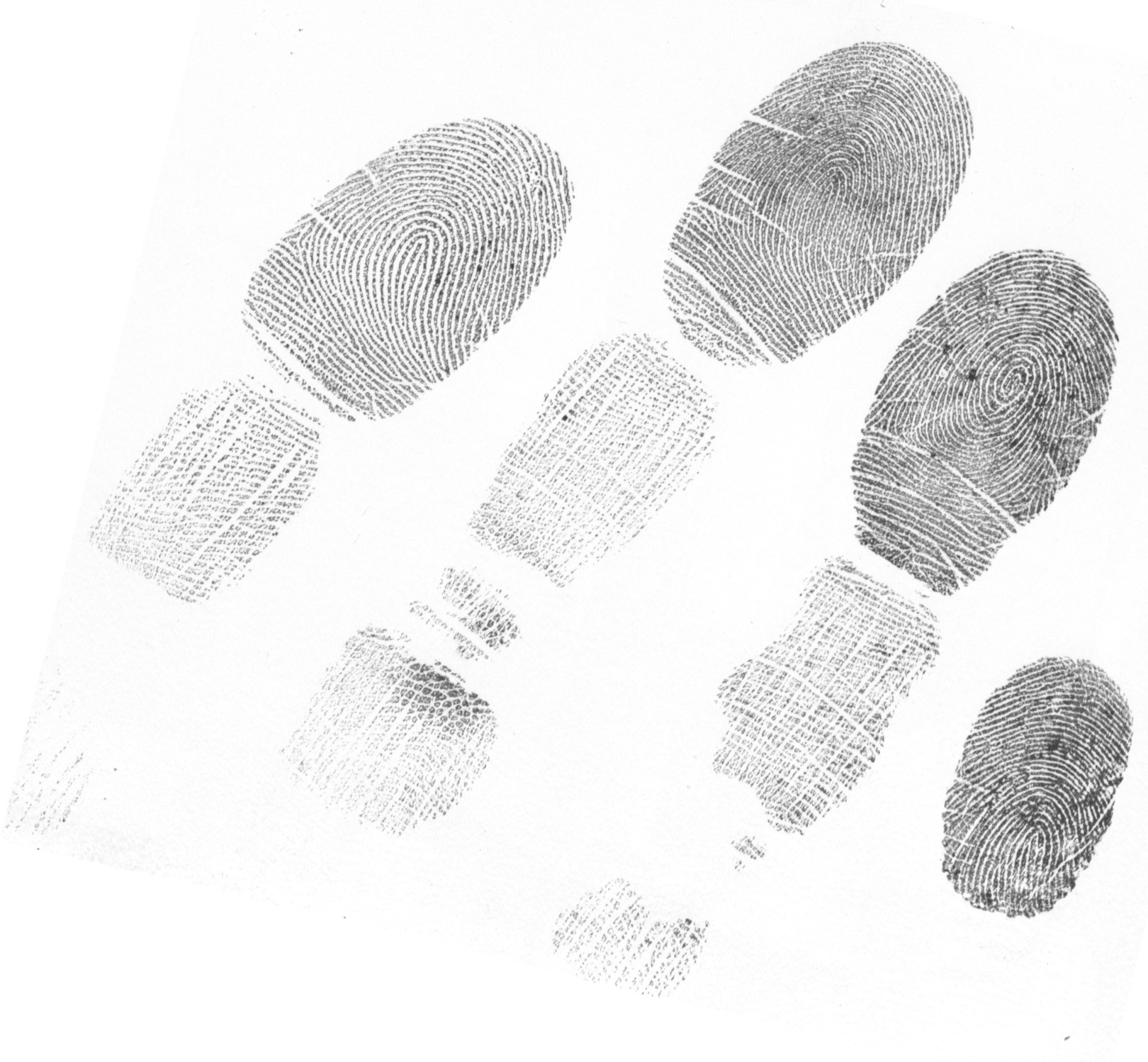}
 \caption{Slap image of the right hand.}
 \label{fig:sampleSlap}    
\end{subfigure}
\begin{subfigure}[b]{0.45\textwidth}
\centering
\begin{subfigure}[b]{0.45\textwidth}
\centering
\includegraphics[width=0.5\linewidth]{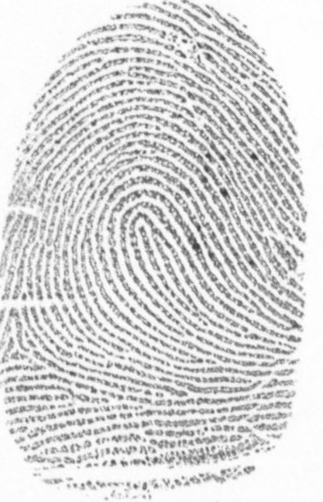}
 \caption{Right index}
 \label{fig:segmented_RightIndexFinger}    
\end{subfigure}
\begin{subfigure}[b]{0.45\textwidth}
\centering
\includegraphics[width=0.5\linewidth]{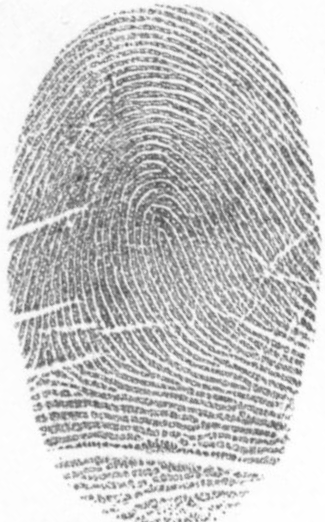}
 \caption{Right middle}
 \label{fig:segmented_RightMiddleFinger}
\end{subfigure}
\begin{subfigure}[b]{0.45\textwidth}
\centering
\includegraphics[width=0.5\linewidth]{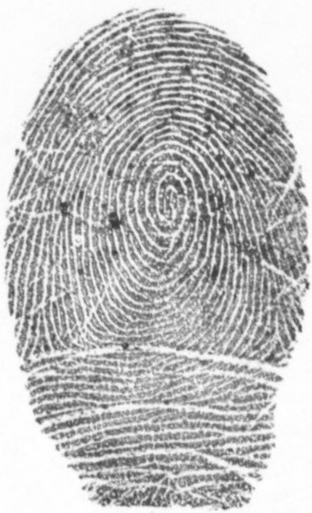}
 \caption{Right ring}
 \label{fig:segmented_RightRingFinger} 
\end{subfigure}
\begin{subfigure}[b]{0.45\textwidth}
\centering
\includegraphics[width=0.5\linewidth]{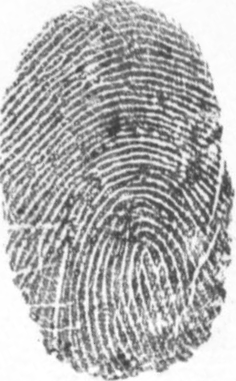}
 \caption{Right little}
 \label{fig:segmented_RightLittleFinger}
\end{subfigure}
\end{subfigure}
\caption{A sample slap image containing four fingerprints of the right hand is shown in \autoref{fig:sampleSlap}. The separation of all fingerprints from a slap image to individual fingerprint images is called slap fingerprint segmentation. The output of segmented images is shown in \autoref{fig:segmented_RightIndexFinger} to \autoref{fig:segmented_RightLittleFinger}.}
\end{figure}

Fingerprints are one of the most used modalities for biometric recognition systems. 
Many applications such as border-crossing, health benefits, and food distribution use slap images instead of single fingerprint images. A slap image contains multiple fingers of a person, usually four fingerprints other than the thumb of the left or right hand, or two fingers of two thumbs. 
\autoref{fig:sampleSlap} shows a right-hand slap containing four fingerprints.  


Previous work on fingerprint-based identification systems suggests higher accuracy when multiple fingers are used instead of a single finger \cite{wilson2004fingerprint}. The first step in a fingerprint identification system with multiple fingerprints is to segment the slap image into N images of individual fingerprints \cite{grother2013biometric};
N is usually four for each of the slap of the left hand and right hand, and two for the thumb slap. Hand-crafted segmentation system NFSEG, created by NIST, is a popular open-source software to segment slap images \cite{Ko2010NBIS}. In 2004 and 2008, the National Institute of Standards and Technology (NIST) organized two contests named SlapSeg04, and SlapsegII to evaluate the performance of segmentation algorithms \cite{Nalla2014DeDuplicationCO}. NFSEG is the only publicly available segmentation algorithm published by NIST. Since then, this algorithm has been used as a benchmark to evaluate the performance of newly developed slap segmentation algorithms \cite{Johnson2010SegmentationOS}. NFSEG has limitations such as poor accuracy and a high failure rate for rotated and in juvenile fingerprints \cite{Murshed2021DeepSlapSeg}.
NIST again conducted a contest called slap fingerprint segmentation evaluation II
(SlapSegII) in 2008. The difference between the two contests is the metrics used for successful
slap fingerprint segmentation .

Recently, Deep Learning (DL) based methods have made significant progress in different computer vision tasks such as object detection, classification, and segmentation \cite{he2017mask}.
Deep learning architectures such as Faster R-CNN, Mask R-CNN, and YOLO exhibit high performance in many aspects of computer vision applications \cite{he2017mask, NIPS2015_14bfa6bb, redmon2016you}. 
Previous work suggests the Slap segmentation system developed using such architectures can potentially outperform NFSEG in terms of fingerprint localization and matching accuracy \cite{Murshed2021DeepSlapSeg}.

Fingerprint matching has become very accurate in recent years \cite{SecurityAnalysis2016Jo}, but many assessments are based on already segmented fingerprints. If a fingerprint is not segmented properly or is mislabeled, it automatically leads to a matching error. Furthermore, due to improvements in fingerprint recognition performance, performance is driven by the edge cases, such as fingers that have extreme rotation, improper orientation, and low quality.


The process of aging can impact the performance of fingerprint identification systems, as changes in skin elasticity and moisture can cause modifications to the patterns and ridges of our fingerprints. In a study conducted by Galbally et al. to investigate the effects of aging on fingerprint biometrics, they found that the accuracy of fingerprint identification systems was more negatively affected when they used fingerprints from children. Specifically, when testing the fingerprints of children aged 0 to 12 years, they reported a decrease of up to 50\% in the accuracy of fingerprint identification systems. 
This leaves a gap in our understanding of how slap fingerprint segmentation algorithms perform in younger populations.
Unfortunately, none of the previously published algorithms for slap segmentation tasks have been evaluated in a dataset containing juvenile and children subjects, as the current benchmark (SLAPSeg competition) only uses data from adult individuals. 

Our approach to solving these problems involves the development of a deep-learning algorithm that can generate both axis-aligned and rotated bounding boxes for precise localization of fingerprints on slap images. To train our proposed algorithm and create a segmentation system called CRFSEG, we utilized a dataset of 11,844 axis-aligned slap images collected from both children and adult subjects that had been manually annotated. Furthermore, we generated arbitrarily angled slap images and their corresponding bounding boxes by rotating axis-aligned slap images and ground-truth bounding boxes from $-90^{\circ}$ to $90^{\circ}$. We constructed a large dataset named \textit{Combined} slap dataset that consists of both axis-aligned and rotated slap images. This dataset is used to train the CRFSEG model in order to efficiently localize fingerprints even at a large angle of rotation. We created another dataset called the \textit{Challenging} slap dataset which consists of challenging images and evaluated the performance of the CRFSEG model on that dataset. More details of our dataset are described in the slap dataset section.


\subsection*{Related Work}
Different types of segmentation methods were used to precisely segment fingerprints and classify them. The backbone algorithms of those methods include convolutional neural network (CNN), Recurrent Adversarial Learning, Fuzzy C-Means and Genetic Algorithm, etc \cite{Murshed2021DeepSlapSeg, 9533712, 9145484, 8852236}. 

A CNN-based fingerprint segmentation technique was proposed by Dai et al., where they consider fingerprint segmentation as a binary classification problem and used a CNN to find background (noise) and foreground (fingerprint) \cite{10.1007/978-3-319-69923-3_35}. This method uses a pre-processing step to remove the background and enhance the ridge structure. After that, the image is divided into different ROIs (fingerprint patches) and fed to a CNN to predict patch categories. To reconstruct the whole fingerprint, a patch quilting method is used after receiving results from the CNN. Finally, morphological operations are used in the contracted segment to obtain a final foreground. Different pre and post-processing techniques are used in this method to get final results. This approach works well when an input image contains one fingerprint. This type of approach will likely fail in slap images because slap images not only capture the fingerprint but also captures other areas under the fingertips as shown in \autoref{fig:sampleSlap}.

Serafim et al. proposed a segmentation technique that finds the regions of interest (ROI) in a fingerprint image using the CNN algorithm and performs patch-level classification of foreground (inside ROI) and background (outside ROI) \cite{8852236}. Their method involves different pre-processing techniques before feeding input images to the CNN and performs different post-processing techniques to obtain the output ROIs. Pre-processing techniques involve dividing an input fingerprint image into different non-overlapping sub-blocks which they called patches. Post-processing involves block filtering, removing islands, pixel filling, and Border smoothing. 

In other research, Stojanovic et al proposed an ROI segmentation method based on deep learning algorithms \cite{7818799}. They used AlexNet and LeNet architectures for detecting ROI in the fingerprints. In this method, the input images are pre-processed which ensures the proper size and acceptable quality of the input image, and then divides the images into multiple non-overlapping patches. Those patches are then fed to the CNN model which generates ROI. After detecting ROIs, region mask margins and different smoothing techniques are used as post-processes methods to get the final results.  

All the methods above rely on pre- and post-processing techniques. A few previous methods can not handle images with different shapes and sizes. Contrasting, our proposed fully automated method can handle images of different sizes and shapes and requires no involvement of humans throughout the entire process.       

\section*{Novelties of this work}
In this study, we focus on developing a deep learning-based age-invariant slap fingerprint segmentation system.  We paid special attention to fingerprints collected from both children and adult subjects, because the shapes, spatial characteristics, and quality of children's fingerprints are different from adult fingerprints \cite{Galbally2019ASO}. This paper describes the following novel contributions:

\begin{itemize}
    \item Built two in-house large datasets named \textit{Combined} and \textit{Challenging} slap datasets that contain 133,611 slap fingerprint images of children and adults.
    \item Annotated all slap fingerprint images manually to establish a ground-truth baseline for the accuracy assessment of different fingerprint segmentation systems.
    \item Developed ``a novel'' age-invariant deep learning-based slap segmentation model (viz. CRFSEG), that can handle arbitrarily orientated fingerprints of adult and juvenile subjects.
    \item Evaluated the performance of state-of-the-art commercial and non-commercial fingerprint segmentation systems using both adult and children slap images and compared the performance of the CRFSEG model with other segmentation systems. 
    \item Released a trained CRFSEG model through GitHub (\url{https://github.com/sarwarmurshed/CRFSEG}) for public use.
\end{itemize}

\section*{Research methods}
In this section, we present in detail the method used for collecting slap image data from children and adult subjects, data annotation, data augmentation, and ground truth creation methods. Then, we describe the proposed neural network architectures. Finally, we discuss the metrics used to evaluate different slap segmentation algorithms.

\subsection*{Slap dataset}
In this section, we discussed slap dataset generation processes including slap collection, annotation, augmentation, and ground truth creation.
\subsubsection*{Slap image collection}
A clean, rich, and representative dataset is what drives complex and sophisticated deep-learning algorithms to deliver state-of-the-art performance and therefore an invaluable resource for building robust deep-learning models. Creating a balanced slap dataset containing fingerprints of both adult and children subjects is challenging since the data need to be captured using special scanners from real subjects.  We collected children's slap fingerprint images from elementary and middle school under an approved IRB.
Collection from children was approved by the Institutional Review Board (IRB) with parents' consent and child asset. We built our novel slap dataset by combining newly collected slap data with several adult slap datasets.   
This dataset contains a total of 15881 slaps (adults: 9131, children: 6750) from 339 adults and 260 children subjects. All slap images are captured at 500 PPI using the FBI-certified \textit{Crossmatch L Scan Guardian (9000251)} fingerprint scanner \cite{noauthor_cross_nodate}.

A rich and representative slap image dataset should contain images that cover a wide variety of scenarios with various poses, illumination, size, brightness, and positions of fingerprints. Such datasets help to build a robust deep-learning-based fingerprint segmentation model and are compiled for a study to robustly evaluate the performance of the segmentation model under real-world conditions.   
After analyzing the data set, we found that many subjects, especially children, put their hands on the scanner at different angles. This results in fingerprints that are often rotated in a slap image. Also, some subjects do not press all their fingers equally, resulting in not all fingers being equally visible in a slap image. Many times, due to the presence of sweat or dust on the hands of a subject, additional data such as lower areas of proximal phalanges which sometimes look like small fingerprints, are added to the slap image. Other common issues in a slap dataset include the presence of the "halo" effect, instances where two neighboring fingerprints touch each other, and amputated or partially captured slaps  \cite{SlapSegII53341}. As a result, the commercial segmentation models that have not been trained with such examples fail in these cases.


\subsubsection*{Slap image annotation and augmentation}\label{subsec:annotationAndBaseline}
Data annotation is an important step in creating a structured and representative dataset which is a prerequisite for building a reliable image segmentation model. Annotation involves drawing a bounding box around each fingertip of a slap image and attaching a label to each fingertip. The fingerprint labels are Left-Index, Left-Middle, Left-Ring, Left-Little, Left-Thumb, Right-Index, Right-Middle, Right-Ring, Right-Little, and Right-Thumb. This annotation is used to train a deep learning algorithm to recognize an area as a distinct object or class in a slap image.

\begin{figure}[!ht]
  \centering
  \includegraphics[width=0.5\linewidth]{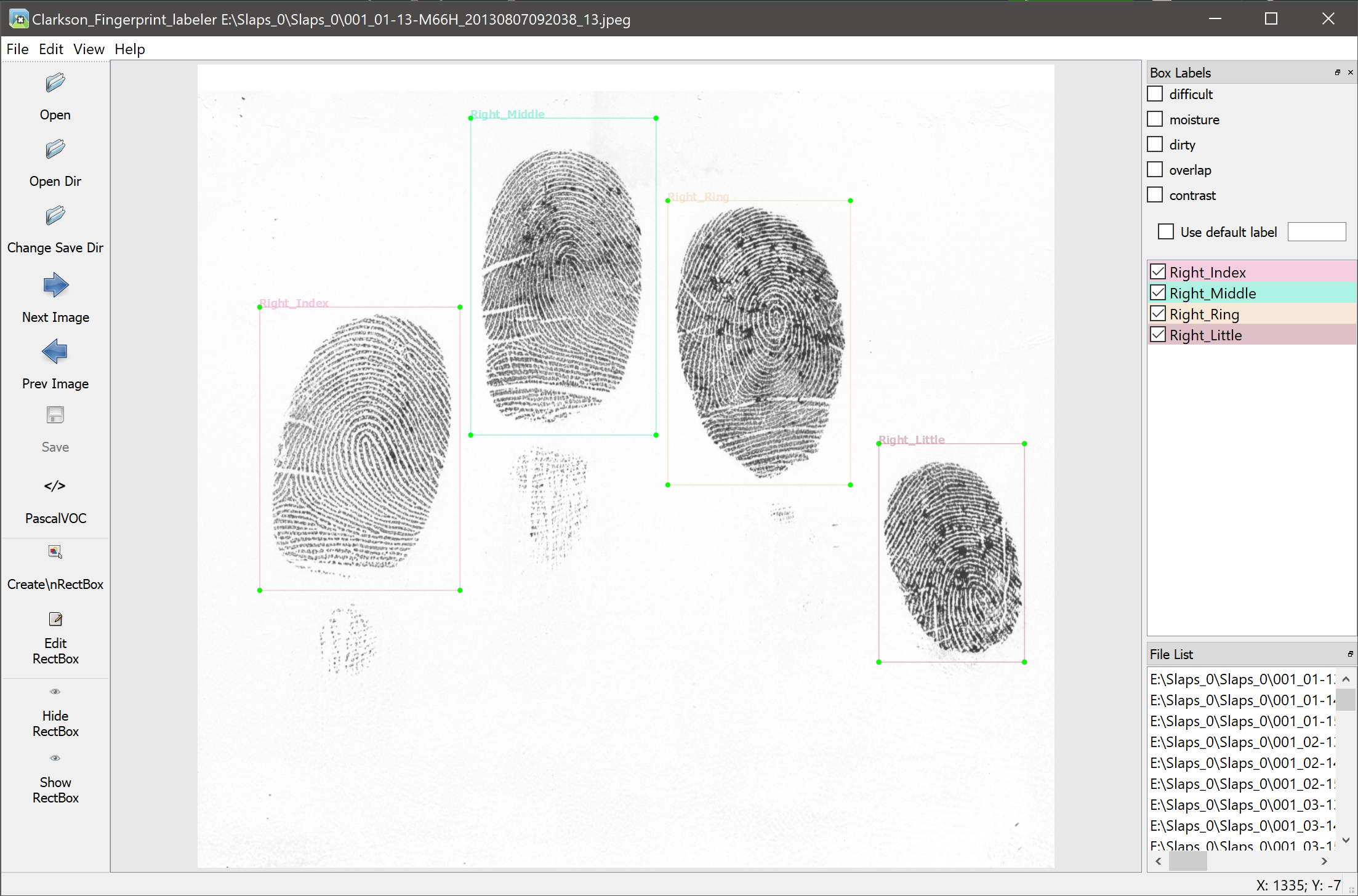}
\caption{Graphical user interface of the slap annotating tool which is built using the Python-based \textit{labelImg} package \cite{Tzutalin}. 
This tool also allows the human annotator to draw, examine and correct the position, orientation, and label of bounding boxes around fingerprints of slap images.}
\label{fig:guiannotation}
\end{figure}

\begin{figure}
\centering
\begin{subfigure}[b]{0.33\textwidth}
\centering
\includegraphics[width=1\textwidth]{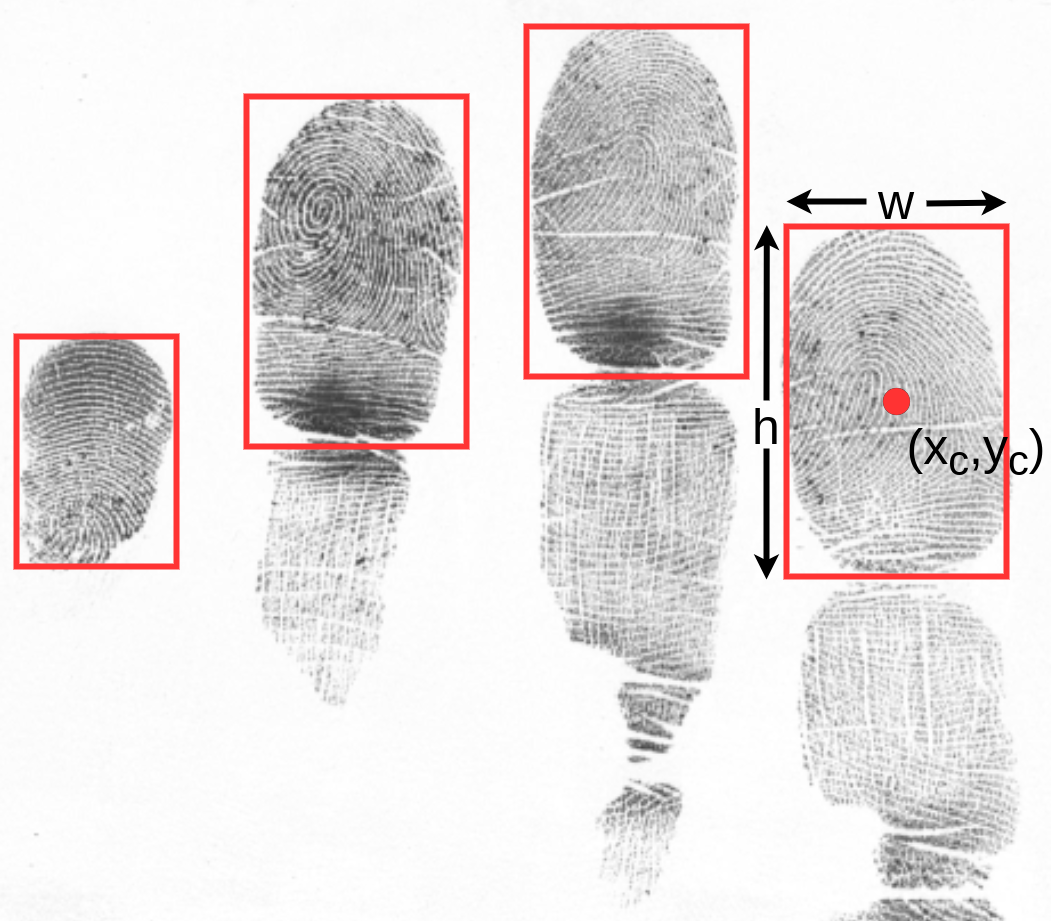}
\caption{Straight bounding boxes around axis-aligned fingerprints.}
\label{fig:straightImgBB}
\end{subfigure}
\hfill
\begin{subfigure}[b]{0.28\textwidth}
\centering
\includegraphics[width=1\textwidth]{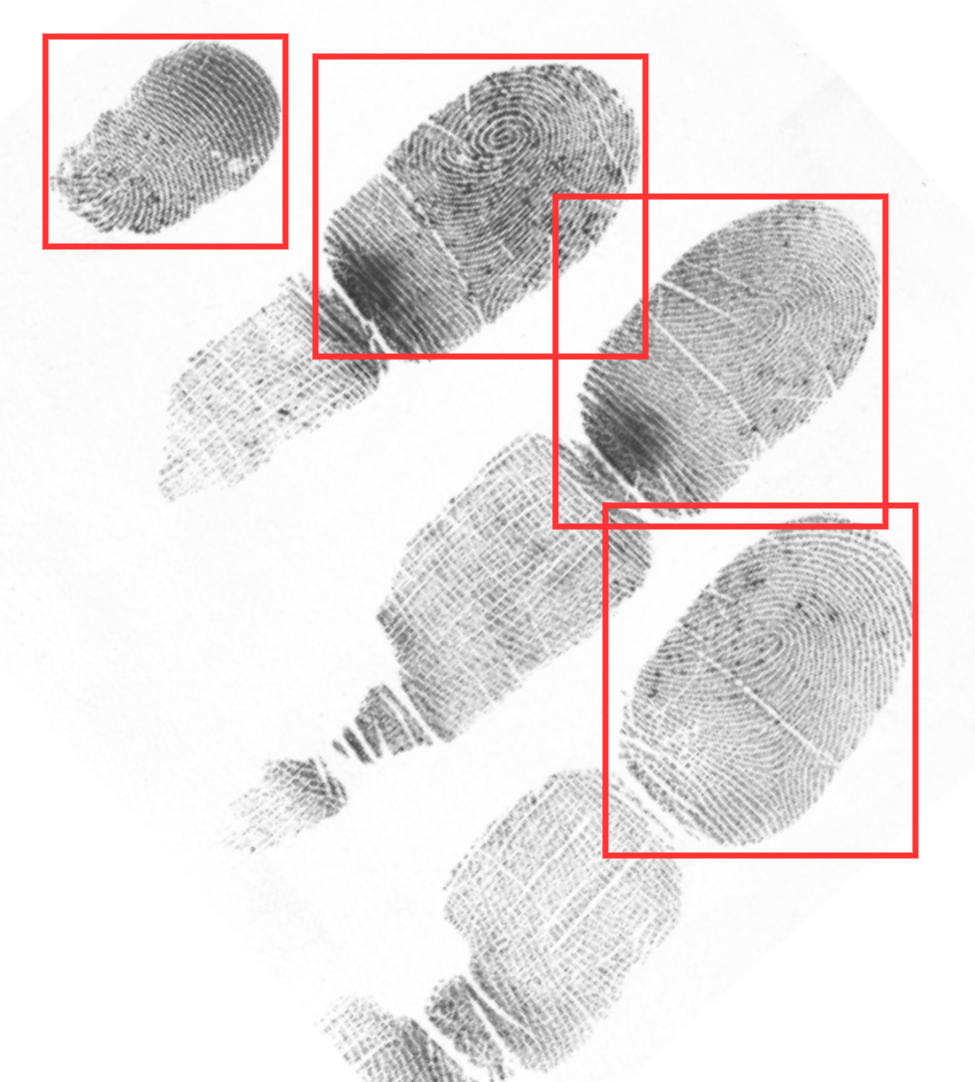}
\caption{Straight bounding boxes around rotated fingerprints.}
\label{fig:straightBBRotatedImg}
\end{subfigure}
\hfill
\begin{subfigure}[b]{0.33\textwidth}
\centering
\includegraphics[width=1\textwidth]{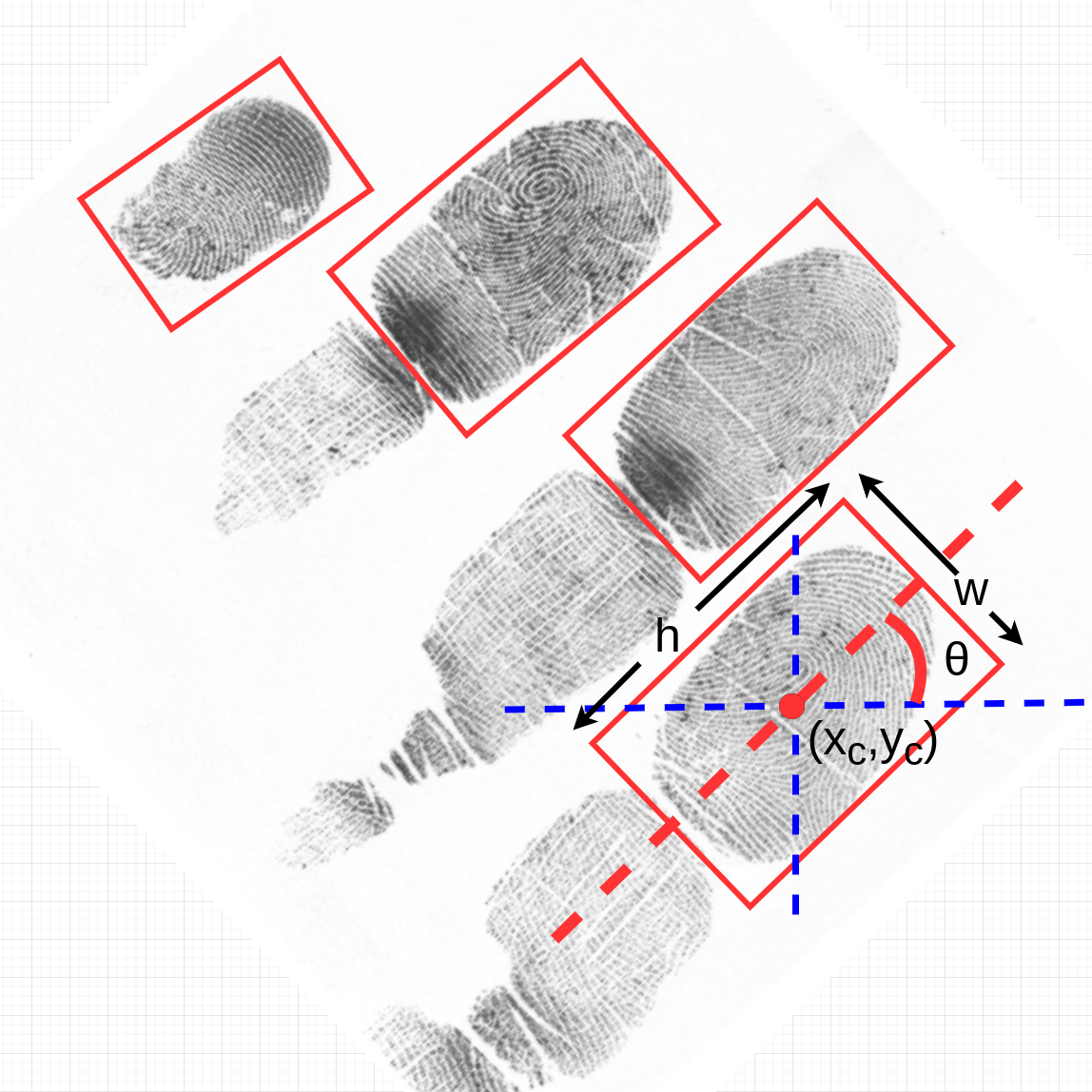}
\caption{Rotated bounding boxes around rotated fingerprints.}
\label{fig:rotatedBBRotatedImg}
\end{subfigure}
 \caption{Three example slap images demonstrating various segmentation strategies. \autoref{fig:straightImgBB} shows four bounding boxes surrounding four axis-aligned fingerprints with good precision. However, when the image is rotated \autoref{fig:straightBBRotatedImg}, the straight bounding boxes are unable of capturing fingerprints precisely, resulting in more areas that sometimes contain regions of neighbor fingerprints. Moreover, some bounding boxes overlap with each other. Capturing more areas with less precise bounding boxes has a negative impact on fingerprint matching.\autoref{fig:rotatedBBRotatedImg} illustrates rotated bounding boxes that enclose targeted fingerprints with high precision, which is an effective segmentation strategy for capturing fingerprints in a rotated slap image.}
 \label{fig:overRotateProblem}
\end{figure}

Manual annotation is the preferred way to create a ground truth dataset for evaluating the performance of the different slap fingerprint segmentation models. However, annotating thousands of images manually is time-consuming and tedious. As an alternative, we leveraged the pre-existing hand information available for all slap images in our dataset, indicating whether the images were from the right or left hand or thumb fingerprints, and utilized NFSEG, a widely used open-source slap fingerprint segmentation model developed by NIST, to segment all images. NFSEG requires hand information to execute the slap segmentation process. 
Slap images that NFSEG failed to segment accurately are labeled as \textit{Difficult} images, while those that were successfully segmented were labeled as \textit{Plain} images.
However, even in the \textit{Plain} images that are successfully segmented by NFSEG, errors may still exist, such as failure to detect all fingerprints on a slap, inaccurate bounding box positions, or incorrect rotation angles. Therefore, we manually reviewed each \textit{Plain} image to ensure the accuracy of each bounding box and corrected any misplaced, mislabeled, or missing bounding boxes for fingerprints. To accomplish this manual inspection task, we used a GUI-based image visualization and annotating software developed using the \verb+labelImg+ \cite{Tzutalin}. 
\autoref{fig:guiannotation} illustrates our fingerprint annotation interface. 
A group of three human annotators examined each \textit{Plain} slap visually using our annotation software, corrected all the errors, and made all the \textit{Plain} slaps straight. 
At the end of this manual examination, we got two types of images, \textit{Plain} images, and \textit{Difficult} images. We then rotated \textit{Plain} images from $-90^{\circ}$ to $90^{\circ}$ to generate arbitrarily angled slap images containing rotated fingerprints.
\begin{enumerate}
    \item \textbf{Plain images}: The slap images where NFSEG was able to correctly segment them, estimate the angle of all slap images and rotate slaps to the upright position. A human annotator cycled through slaps and corrected the position of bounding boxes around each finger that was incorrectly positioned and labeled by NFSEG, while the remaining bounding box(es) remained untouched. This process resulted in a cleaned and annotated 11844 slap images with a total of 40221 localized fingerprints (adult: 24234, children: 15987). All the \textit{Plain} images were used for data augmentation. The distribution details of the \textit{Plain} images can be found in \autoref{table:datasetdistribution}.

    \begin{table}[t]
    \renewcommand{\arraystretch}{1}
    \caption{Description of each slap fingerprint dataset used for training and testing.
    The \textit{Plain} dataset contains images that are successfully segmented by the state-of-the-art NIST NFSEG, while the \textit{Challenging} dataset includes images that NFSEG failed to segment. Data augmentation techniques are applied to \textit{Plain} annotated images to synthetically generated rotated images containing over-rotated fingerprints. The \textit{Plain} and rotated images are mixed to build a \textit{Combined} dataset containing both straight and over-rotated images. This \textit{Combined} dataset is used to build the deep learning-based slap fingerprint segmentation model.}
    \label{table:datasetdistribution} 
    \centering
    \begin{tabular}
    {|c|c|c|c|c|c|c|}
    \hline 

    {Dataset} & {Image types} & {Age groups} & {Total slaps} & {Left hand slaps} & {Right hand slaps} & {Thumb slaps} \\
    \hline
    \hline 
     \multirow {4}{*}{\centering \textit{Combined}} & \multirow {2}{*}{\parbox{3.5cm} {\textit{Plain}: Successfully segmented by the NFSEG}}  & Children & 4642 & 1341  & 2075 & 1226 \\ \cline{3-7}
    {} & {} & Adult & 7202 & 2148  & 2811 & 2243 \\
    \cline{2-7}
    {} & \multirow {2}{*}{\parbox{3.5cm} {\textit{Augmented}: Rotated images of \textit{Plain} dataset}}  & Children & 45960 & 13190  & 20660 & 12110 \\ \cline{3-7}
    {} & {} & Adult & 71770 & 21340  & 28070 & 22360 \\
    \hline
    \multirow {2}{*}{\centering \textit{Challenging} } & \multirow {2}{*}{\parbox{3.5cm} {\textit{Difficult}: Failed to segment by the NFSEG}}  & Children & 2108 & 925  & 169 & 1014 \\ \cline{3-7}
    {} & {} & Adult & 1929 & 909  & 229 & 791 \\
    \hline
    \end{tabular}
    \end{table}

    \item \textbf{Difficult images}: A significant portion, around 25.4\%, of the slap images in the dataset were not properly segmented by the NFSEG due to the inability to determine accurate rotational angles. These images were categorized as \textit{Difficult} images. 
    Upon manual examination, we found that NFSEG failed to segment an image when that image has fingers with weak and low resolution due to noise, deviation, variation, and particularly when fingerprints on a slap image are rotated more than $ \pm 35^{\circ}$ from the upright position. The total number of \textit{Difficult} slap images is 4037 containing 12370 fingerprints (adult: 6071, children: 6299). The \textit{Difficult} slap images were not annotated manually, and therefore, they were not used for training or validating the deep learning model. Instead, these images were exclusively utilized for assessing different slap segmentation systems. Details of the \textit{Difficult} images are shown in \autoref{table:datasetdistribution}.

    \item \textbf{Augmented images}: The NFSEG segmentation model fails to segment slaps when it fails to calculate the correct rotation angle of the slap. NFSEG predicts out-of-position bounding boxes around the fingerprints for most of these failure cases. Sometimes, predicted bounding boxes overlap each other. Example failure cases of the NFSEG model are shown in \autoref{fig:FailedByNFSEG}. 
    
    \begin{figure}
    \centering
    \begin{subfigure}[b]{0.33\textwidth}
    \centering
    \includegraphics[width=1\textwidth]{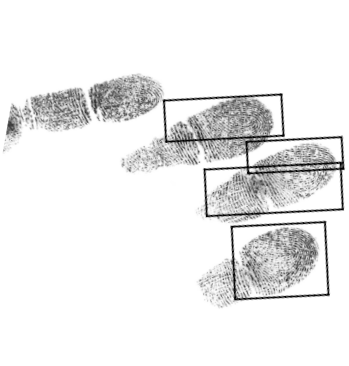}
    \end{subfigure}
    \hfill
    \begin{subfigure}[b]{0.33\textwidth}
    \centering
    \includegraphics[width=1\textwidth]{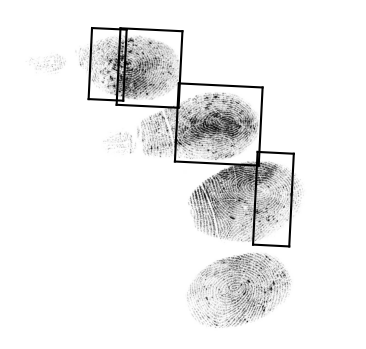}
    \end{subfigure}
    \hfill
    \begin{subfigure}[b]{0.33\textwidth}
    \centering
    \includegraphics[width=1\textwidth]{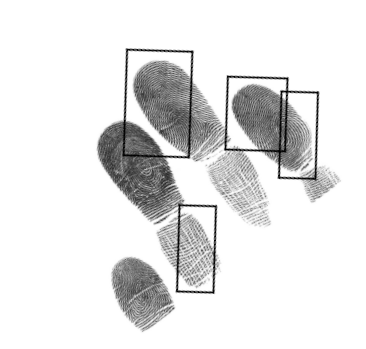}
    \end{subfigure}
     \caption{Examples of failed segmentation by NFSEG. This model predicted incorrect positions of bounding boxes if the slap images contained fingerprints that are rotated, i.e., not axis-aligned. These images are taken from the \textit{Combined} dataset. We observed that the number of failures was higher in slaps collected from children subjects.}
     \label{fig:FailedByNFSEG}
    \end{figure}
    
    Additionally, the failure rate of the NFSEG model increases if the slap is not axis-aligned but rotated. VeriFinger segmentation software returns a failure message such as "TooFewObjects" or "BadObject" instead of returning the incorrect positions and incorrect labels of bounding boxes \cite{VeriFingerSDK}. Our previously developed slap segmentation model CFSEG \cite{Murshed2021DeepSlapSeg} generates axis-aligned straight bounding boxes with no rotational angles of the predicted bounding boxes resulting in the overlap between bounding boxes when a slap is rotated. \autoref{fig:straightBBRotatedImg} shows a failure case by our previous segmentation model. To overcome the aforementioned segmentation failures in over-rotated slap images, we developed a deep learning-based robust slap segmentation model named CRFSEG model that predicts the angle of each fingerprint in a slap image individually. Training such models requires a dataset that contains both rotated and straight slap images. 
    A minimum of four parameters are required to locate a fingerprint in a slap image using a bounding box. 
    These parameters are ($x_c, y_c, w, h$), where $(x_c, y_c)$  represents the center coordinates, $w$ the width and $h$ the height of a horizontal bounding box around a fingerprint as shown in \autoref{fig:straightImgBB}. 
    However, horizontal bounding boxes do not describe the fingerprint outline with high precision when the fingerprints are tilted or rotated relative to the horizontal axis which can happen when the full slap has not been axis aligned or individual fingers are rotated relative to each other. 
    \autoref{fig:straightBBRotatedImg} illustrates such a case, where some bounding boxes not only capture more areas than the actual fingerprint but also overlap each other. Capturing such areas reduces the preciseness of the segmentation capability of the slap fingerprint segmentation model, resulting in fingerprint-matching failure. 
    An additional parameter is needed to accurately locate a rotated fingerprint and to reduce the unwanted area captured by the bounding boxes. 
    Therefore, along with four original parameters, we added a new parameter $\theta$ to represent a rotated bounding box ($x_c, y_c, w, h, \theta$), where, ($x_c, y_c$) is the center of the bounding box, $w$ and $h$ represent the width and height of the bounding box respective, and $\theta$ represents the angle (range of [$-90^\circ$ to  $90^\circ$]) of the bounding box relative to the vertical axis. All parameters and rotated bounding boxes are shown in \autoref{fig:rotatedBBRotatedImg}. 
    
    The \textit{Plain} image set contains a total of 11844 annotated slap images that are oriented upright with respect to the y-axis. An image of such upright oriented slap is shown in \autoref{fig:straightImgBB}.  
    The data augmentation techniques are applied to all \textit{Plain} annotated images and generated augmented images containing over-rotated fingerprints. Random rotations from $-90^\circ$ to  $90^\circ$ are applied to the \textit{Plain} images to generate those rotated images. A total of 117730 rotated slap images are generated using augmentation techniques. The distribution details of the augmented images are shown in \autoref{table:datasetdistribution}. 
\end{enumerate}

\subsubsection*{Dataset naming and ground truth}
We mixed \textit{Plain} and augmented images together and made a \textit{Combined} dataset that contains both upright and rotated slap images. A total of 129,574 annotated images are included in this dataset, which was subsequently used as the ground truth to test different segmentation systems.  We also created a separate dataset called the \textit{Challenging} dataset, consisting of images that were labeled as \textit{Difficult} images. The distribution details of the \textit{Combined} and \textit{Challenging} datasets are presented in \autoref{table:datasetdistribution}. We used only the \textit{Combined} dataset for training, validation, and testing of the CRFSEG model, and compared its performance with that of NFSEG and VeriFinger (a commercial slap segmentation software \cite{VeriFingerSDK}) in the results section. 

After training and validating the CRFSEG model on \textit{Combined} dataset, we evaluated its performance on the \textit{Challenging} dataset. We also evaluated VeriFinger on the the same \textit{Challenging} dataset and compared its performance with the CRFSEG. The comparison results were reported in the results section.

\subsection*{Proposed deep learning-based network architecture}
Fingerprints of adult and juvenile subjects have different shapes and sizes. To handle the size variations of fingerprints, we utilized a two-stage Faster R-CNN architecture consisting of a Feature Pyramid Network (FPN) which helps to detect objects at different scales \cite{NIPS2015_14bfa6bb}. 
The original Faster R-RCNN architecture is designed to detect objects with high accuracy and draw a horizontal axis-aligned bounding box around the detected object. 
However, this horizontal bounding box suffers several problems including capturing unnecessary areas, overlapping with close adjacent boxes, and less precise object localization. 
To overcome such issues, we modified the original architecture of Faster R-CNN and included new layers to generate rotated bounding boxes around fingerprints. 
We named this new architecture the Clarkson Rotated Fingerprint segmentation (CRFSEG) system. 
This architecture has three building blocks namely, 
\begin{enumerate*}[label*=(\roman*)]
    \item Backbone network,
    \item Oriented region proposal network, and
    \item Box head.
\end{enumerate*}

\subsubsection*{Backbone network}
The backbone network of CRFSEG is a ResNet-50 with FPN architecture that extracts feature maps from input slap images at different scales. 
The main components of ResNet-50 are a stem block and multiple bottleneck blocks. The stem block contains three main layers including two-dimensional convolution layers (Conv2D), rectified linear unit (ReLU), and two-dimensional max pooling layers. 
The Conv2D down-samples the input image twice by running the convolution with kernel size = 7, stride = 2, and max pooling with stride = 2. 
This type of steam block reduces the information loss when an input image is processed through the network, resulting in improved network expression ability without increasing the computational cost.
On the other hand, bottleneck blocks, which are originally proposed in the ResNet paper, are used for efficient computation \cite{He2016RESNET}. The bottleneck block has three main convolution layers with convolution kernel sizes of 1$\times$1, 3$\times$3, and 1$\times$1 respectively. The first 1$\times$1 is used to  decrease the number of calculations required by decreasing the number of input channels. The next block with 3$\times$3 convolution kernel is deployed to extract features for the network. The final block with 1$\times$1 block increases the channels. The output of the backbone network is multi-scale semantic-rich feature maps which are used as input to the oriented region proposal network (O-RPN).

\subsubsection*{Oriented Region Proposal Network (O-RPN)}
After obtaining the semantic-rich feature maps of the input image from the backbone network, the Oriented Region Proposal Network (O-RPN), which is a small network, uses those feature maps to propose the oriented regions of interest (ROIs) where the probability of having a targeted object is high. 
Proposing ROIs helps to generate final rotated bounding boxes resulting in precise bounding boxes for both axis-aligned and rotated objects. This type of regional proposal network is different from the conventional regional proposal network (RPN) used in the Faster R-CNN architecture where only axis-align regions are proposed by the RPN. 

\begin{figure}
 \includegraphics[width=1\linewidth]{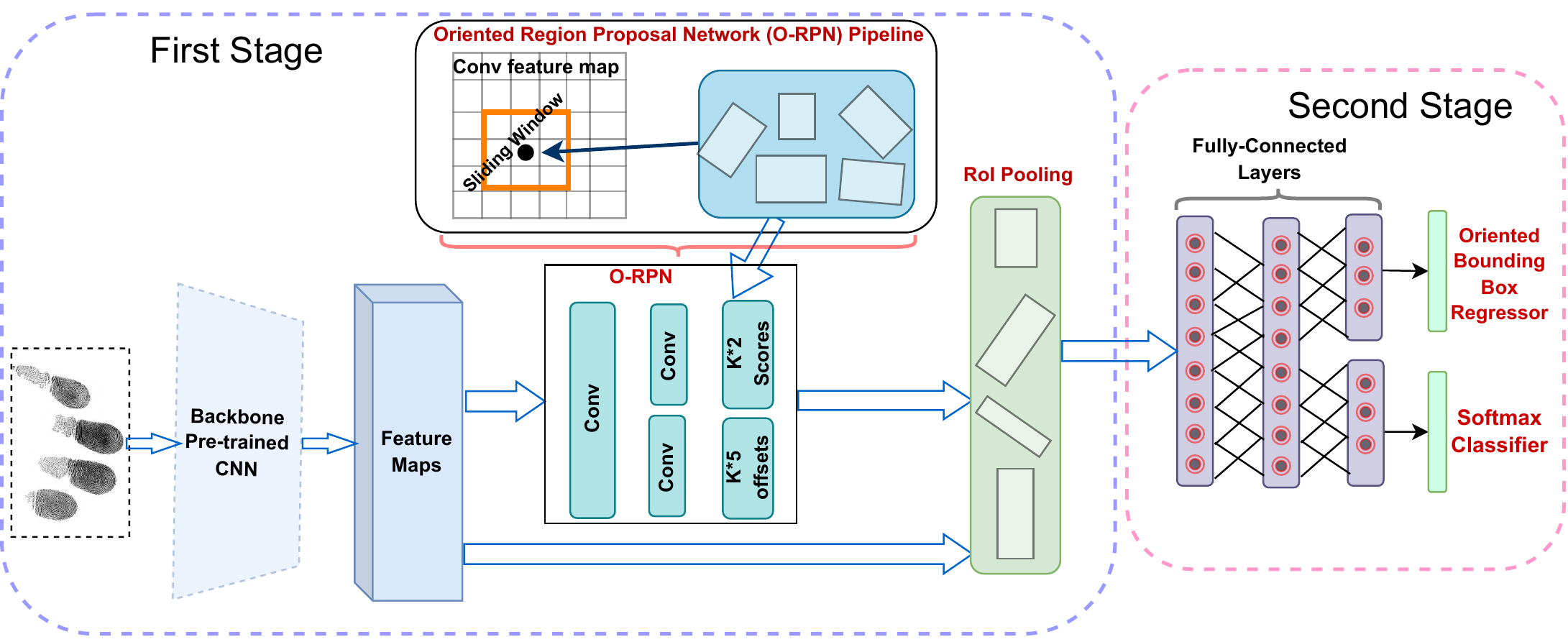}
 \caption{The complete architecture of Clarkson Rotated Fingerprint Segmentation (CRFSEG) system. An input fingerprint image is processed by a backbone CNN model that is pre-trained on the ImageNet dataset and generates feature maps. Then, O-RPN operates on all levels of the feature maps and generates oriented anchors. The RoI Pooling layer generates fixed-length feature vectors by selecting spatial features from the output of the backbone network and O-RPN. These fixed-length feature vectors are then passed through a sequence of fully connected layers. The output of fully-connected layers is fed to two parallel branches. One branch is named Oriented Bounding Box Regressor which contains regressors for bounding box regression and the other is named Softmax Classifier which contains softmax layers for multiclass classification.}
 \label{fig:rotatedFasterRCNNArchi}
\end{figure}

To propose the oriented regions, a 3$\times$3 sliding window runs through the feature maps and generates a set of boxes referred to as anchors with different orientations, scales, and aspect ratios. If we have $k_a$ different orientations, $k_s$ different scales, and $k_r$ different aspect rations, then  $K = k_a \times k_s \times k_r $ numbers of anchor boxes are generated for each position in the feature map. Most of these anchor boxes might not have targeted objects in them. A sequence of convolution layers and two parallel output layers, which we call localizing and classifying layers, are used to separate anchors containing target objects from other anchors. The classifying layer calculates the intersection-over-union (IoU) score of the ground truth box with anchor boxes and classifies the anchors as foreground that contains targeted objects, or background that contains no objects. The localizing layer learns the offsets (x,y,w,h, $\theta$) values for the foreground boxes. 
A (K$\times$5) parameterized encodings for regression offset are generated by the localizing layer and (K$\times$2) parameterized scores for region classification are generated by the classifying layer. The anchor generation strategy is illustrated in \autoref{fig:rotatedFasterRCNNArchi}.

For training O-RPN, seven different orientations (-$\pi/4$, -$\pi$/6, -$\pi$/12, 0, $\pi$/12, $\pi$/6, $\pi$/4), three different aspect ratios (1:1, 1:2, 2:1 ) and three different scales (128, 256 and 512) are used to generate anchors. Then, all the anchors are divided into three categories 
\begin{enumerate*}[label*=(\roman*)]
    \item positive anchors: anchors that have greater IoU overlap or an IoU overlap larger than 0.7 with respect to the ground-truth boxes, 
    \item negative anchors: anchors that have an IoU overlap with a ground-truth box smaller than 0.3,
    \item neutral anchors: anchors that have IoU overlap between 0.3 and 0.7 with respect to ground-truth boxes. These types of anchors are removed from the anchor set and are not used during training.
\end{enumerate*}
Note that, unlike Faster R-CNN, which uses horizontal anchor boxes during the calculation of IoU overlaps, we use oriented anchor boxes and oriented ground-truth boxes.
The equations below show the loss functions that are used to train the O-RPN. 
\begin{equation}
    L_{o-rpn} = L_{cls}(p,u) + \lambda u L_{reg}(t, t^*) 
\end{equation}
Here, $L_{cls}$ is the classification loss, $p$ is the predicted probability over the foreground and background classes by the softmax function, $u$ represents the class label for anchors, where u = 1 for foreground containing fingerprint and u = 0 for background; $t = (t_x, t_y, t_h, t_w, t_\theta)$ denotes the predicted regression offset value of an anchor calculated by the network, and $t^* = (t^*_x, t^*_y, t^*_h, t^*_w, t^*_\theta)$ denotes the ground truth. $\lambda$ is a balancing parameter that controls the trade-off between class loss and regression loss. The regression loss is activated only if u = 1 for the foreground and there is no regression for the background.

We define the classification loss function as cross-entropy loss between the ground-truth label $u$ and predicted probability $p$ as:
\begin{equation}
    L_{cls}(p,u) = -u \cdot \log(p) - (1-u) \cdot \log(1-p)
\end{equation}
The tuple $t$ and $t^*$ are calculated as follows:
\begin{equation}
    t_x = (x - x_a)/w_a, t_y = (y-y_a)/h_a, \\
    t_w = \log (w/w_a), t_h = \log(h/h_a),\\
    t_\theta = \theta-\theta_a
\end{equation}
\begin{equation}
    t^*_x = (x^* - x_a)/w_a, t^*_y = (y^*-y_a)/h_a,\\ 
    t^*_w = \log (w^*/w_a), t^*_h = \log(h^*/h_a),\\
    t^*_\theta = \theta^*-\theta_a
\end{equation}
Where, where $x$, $x_a$ and $x^*$ denote the predicted box, anchor and
ground truth box, respectively; the same is for $y$, $h$, $w$ and $\theta$.
The smooth-L1 loss is adopted for the bounding box regression as follows:
\begin{equation}
    L_{reg}(t, t*) = \sum_{i \in {x,y,w,h,\theta}} u.\text{smooth}_{L1} (t^*_i - t_i)
\end{equation}
\begin{equation}
    \text{smooth}_{L1} (x) = \left\{
    \begin{array}{ll}
        {0.5x^2} & \mbox  {if |x| <1} \\
        {|x| - 0.5} & \mbox {\text{otherwise}}
    \end{array}
\right. 
\end{equation}

\subsubsection*{Box head}
By default, the O-RPN network generates 1000 proposal boxes and objectness logits. 
An ROI pooling layer is used to project proposal boxes into feature space and the output of this layer is reshaped and fed to the fully connected (FC) layers. Then, the RoI vector is generated by the FC layers and passed through a predictor containing two branches named rotated bounding box regressor and classifier. Finally, the classification layer of the model predicts the object class, and the regressor layer regresses bounding box values.  

\subsection*{Evaluation metrics}
We used Mean Absolute Error (MAE), fingerprint angle prediction error, fingerprint labeling accuracy, and matching score to evaluate and compare the performance of the CRFSEG model with the NIST fingerprint segmentation system, NFSEG, and a commercially used fingerprint segmentation software named Neurotechnology VeriFinger \cite{VeriFingerSDK}. 

\begin{enumerate}
    \item {Mean Absolute Error (MAE)}: The MAE measures the capabilities of slap segmentation systems to correctly segment fingerprints within a certain geometric tolerance of human-annotated ground truth data. The minimum Geometric Tolerance Limit (GTL) allowed in NIST Slapseg-II is -32 pixels for the left and right sides, -64 pixels for the top, and bottom sides \cite{watson_slapsegii_2010}.  
    Slap fingerprint segmentation models have to find the balance between over-segmentation (small bounding box covering less area than the ground truth fingerprint area) vs. under-segmentation (over-extending the predicted bounding box covering more area than the ground truth fingerprint area). 
    Over-segmentation can lead to capturing the ridge-valley structure of close adjacent fingerprints during segmentation. 
    This extra noise can potentially degrade the matching performance. On the other hand, under-segmentation can result in the loss of valuable parts of the fingerprints leading to the degradation of matching performance. 
    MAE is a metric that is used to determine over-segmentation or under-segmentation by the model. 
    
    To measure the MAE of a detected bounding box, we calculated the distance of each side of a detected bounding box from the corresponding side of the annotated ground-truth bounding box in terms of pixels. \cite{Murshed2021DeepSlapSeg}. 
    A successful segmentation refers to finding a bounding box around a fingerprint within a certain geometric tolerance of the human-annotated ground truth bounding box.
    \begin{figure}
    \centering
    \includegraphics[width=0.4\linewidth]{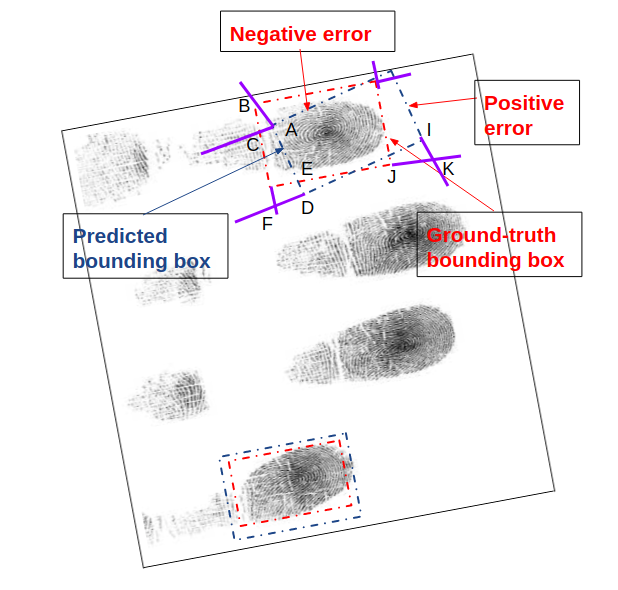}
     \caption{Example for calculating a positive and a negative error between the predicted and ground truth bounding box using Euclidean distance. To calculate the pixel error of each side of the predicted bounding boxes, we first calculate the distance of the endpoints of a side from the corresponding endpoints of the annotated ground-truth bounding box in terms of pixels. For instance, during calculating the pixel error of the side $AD$, we draw two perpendicular lines $AC$ and $DF$ with the line $AD$. The perpendicular line $AC$ intersects the corresponding ground-truth line at point $C$. Another perpendicular line $DF$ intersects the extension of the corresponding ground-truth line at point $F$. Finally, the Euclidean distances from point $A$ to point $C$  and point $D$ to point $F$ are calculated. The average of these two Euclidean distances is the pixel error for the side $AD$. We apply the same approaches to calculate all four sides of a bounding box. Finally, \autoref{equ:MAEs} is applied separately on pixel errors of each size to calculate the MAE of that side.}
     \label{fig:MAECalculation}
    \end{figure}
    
    In our previous study, we measured the MAE for the bounding boxes that are axis-aligned (straight from top to bottom) \cite{Murshed2021DeepSlapSeg}. However, in this study, we need to measure MAE for rotated bounding boxes. For measuring the exact pixel loss, we separately measure the average Euclidean distance for all four sides of the detected bounding box. To measure the Euclidean distance of any side of a detected bounding box, first, we draw two perpendicular lines from two distance points of a side and find two intersect points of those perpendicular lines and the corresponding side of the ground-truth bounding box. If necessary we extend the length of the ground-truth side so that it intersects with the line drawn perpendicular to the detected side. 
    \autoref{fig:MAECalculation} illustrates more detail about calculating the MAE for a fingerprint. If any side of a detected bounding box captures more information than the side of the ground-truth bounding box, we consider it as a positive error. An example of a positive error is shown on the right side of the rightmost fingerprint in  \autoref{fig:MAECalculation}. If any side of a detected bounding box captures less information than the ground-truth bounding box, we consider it as a negative error. An example of the negative error is shown in the top side of the left-most fingerprint in \autoref{fig:MAECalculation}.
    
    
    Finally, the MAE for each side is calculated using \autoref{equ:MAEs} separately.    
    \begin{equation}
    \label{equ:MAEs}
        \text{MAE} = \frac{1}{N}{\sum_{i = 0} ^ {N} |X \: error_i|}
    \end{equation}
    Where $N$ is the total number of fingerprints in the test dataset. $X$ represents the Euclidean distance error of any side such as left, right, top, or bottom of the bounding box.

    \item Error in Angle Prediction (EAP): To evaluate the capability of different fingerprint segmentation systems to predict the orientation of fingerprints, we calculate the deviation of the predicted angle compared to the ground truth using \autoref{equ:EAP}.
    \begin{equation}
    \label{equ:EAP}
        \text{EAP} = \frac{1}N{\sum_{i = 0} ^ {N} |\theta - \theta^*|}
    \end{equation}
    Where $N$ is the total number of fingerprints in the test dataset. $\theta$ is a ground truth angle and $\theta^*$ is a predicted angle by a fingerprint segmentation model. 
    
    \item Fingerprint classification accuracy: Hamming loss is used to evaluate the performance of multi-class classifiers \cite{Tsoumakas2007MultiLabelCA}. Hamming loss measures the number of positions in which the corresponding symbols are different between two equal-length sets: a set of the ground-truth label and a set of the predicted label.  \autoref{equ:HLequation} is used to calculate the Hamming loss \cite{Tsoumakas2007MultiLabelCA}.
    \begin{equation}
        \label{equ:HLequation}
        \text{Hamming loss} = \frac{1}{N} \sum_{i = 1} ^ {N} \frac{|Y_i \Delta Z_i|}{|L|}
    \end{equation}
    where $N$ is the total number of samples in the dataset, and $L$ is the number of labels. $Z_i$ is the predicted value for the i-th label of a given sample, and $Y_i$ is the corresponding ground true value. $\Delta$ stands for the symmetric difference between two sets of predicted and ground true value.

    The accuracy of a multi-class classifier is related to Hamming loss \cite{ha2021topic, koyejo2015consistent}, which can be calculated using \autoref{equ:AccuracyHL}. 
    \begin{equation}
    \label{equ:AccuracyHL}
        \text{Accuracy} = 1 - \text{Hamming loss}
    \end{equation}
    
    \item Fingerprint Matching: This is one of the most important aspects of slap segmentation systems.
    Fingerprint matching is performed to measure the capabilities of slap segmentation algorithms to correctly segment fingerprints within a certain tolerance.
    We use the true accept rate (TAR) and false accept rate (FAR) to report fingerprint matching. 
    TAR is the percentage of instances a biometric recognition system verifies an authorized person correctly, which is calculated using \autoref{equ:TAR}:
    \begin{equation}
    \label{equ:TAR}
        TAR = \frac{\text{Correct accepted fingerprints}}{\text{Total number of mated matching attempts}} \times 100\%
    \end{equation}
    
    FAR is the percentage of instances that a biometric recognition system verifies an unauthorized user, calculated using \autoref{equ:FAR}.
    \begin{equation} 
    \label{equ:FAR}
        FAR = \frac{\text{Wrongly accepted fingerprints}}{\text{Total number of non-mated matching attempts}} \times 100\%
    \end{equation}

\end{enumerate}

\section*{Experiments and results}
In this section, we provide details of the experiments conducted to measure the current capabilities of different slap segmentation algorithms and demonstrate the effectiveness of our newly developed CRFSEG model. 
We describe the dataset used in our experiments and the training process of CRFSEG. 
We present the comparisons with existing segmentation algorithms and demonstrate that CRFSEG outperformed state-of-the-art segmentation systems in terms of segmentation, finger labeling, as well as matching accuracy.

\subsection*{Distribution of training dataset}

We use the split ratio of 80:10:10 of the \textit{Combined} dataset, where 80\% slap images are used for training the deep learning model, 10\% slap images are used for validating the model, and the rest of the 10\% images are used for testing the model. We used the 10-fold cross-validation technique to build the model and evaluate its performance. All evaluation results are reported in the results section.

After building a robust segmentation model (CRFSEG), we used the separate \textit{Challenging} dataset to test the performance of this deep-learning-based model. We also evaluate the performance of the VeriFinger slap segmentation software using this \textit{Challenging} dataset.
We observed that the VeriFinger slap segmentation software struggled to segment slap images in the \textit{Challenging} dataset. On the contrary, our newly developed CRFSEG model performed better on that dataset and outperformed the VeriFinger segmentation system by achieving a 10\% higher matching score.  

\subsection*{Training}
CRFSEG model was developed based on the Detectron2 software system developed by Facebook AI Research (FAIR) \cite{wu2019detectron2}. Detectron2 supports different state-of-the-art deep learning-based object detection algorithms such as Faster R-CNN, RetinaNet, etc. We adopted the Faster R-CNN algorithm and modified the code of Detectron2 software to implement the oriented regional proposal network (ORPN), to add new layers to the regressor in order to find the offset of the rotated bounding boxes and to incorporate all our requirements to segment fingerprints from a slap image accurately. 
The Faster R-CNN model used in our experiment was pre-trained on the MS-COCO dataset where the number of output classes is 81. We changed the output layers and reduced the class number from 81 to 10 in order to classify ten fingerprints from two hands. After that, the model was slowly fine-tuned using our novel slap image dataset.  
 
We used the end-to-end training strategy, where loss values were obtained by comparing the predicted result with the true result. Training was done on a Linux-operated desktop machine with 20 cores Intel(R) Xeon(R) E5-2690 v2 @ 3.00GHz CPU, 64 GB RAM, and with a NVIDIA GeForce 1080 Ti 12-GB GPU. 
A total of 28000 iterations was used to train the fingerprint segmentation model with learning rates starting from $10^{-3}$, and are multiplied by $0.1$ after 4000, 8000, 12000, 18000, and 25000 iterations. Weight decays are 0.0005, and momentums are 0.9. All experiments used multi-scale training, hence we did not need to scale up or down any input before feeding it to the neural network. 

\subsection*{Results}
\subsubsection*{Mean Absolute Error (MAE) for segmented bounding boxes}
MAE measures the preciseness of bounding boxes around fingerprints generated by slap segmentation algorithms.
We used \autoref{equ:MAEs} to calculate the MAE of different segmentation algorithms.
\autoref{table:AvgMAE} shows the MAE and its standard deviation for NFSEG, VeriFinger, and CRFSEG models calculated on the \textit{Combined} slap dataset. 
The MAEs of the CRFSEG model in prediction bounding boxes on both adult and children subjects are significantly lower, which indicates less pixel error and more precise bounding boxes, compared to NFSEG and VeriFinger segmentation software. 
Additionally, our results confirm that compared to adults, NFSEG achieves lower performance for the children subjects in \textit{Combined} dataset.
Furthermore, NFSEG struggles to localize the bounding boxes when slap images are not axis-aligned but rotated. 
Compared to NFSEG, VeriFinger performs better, 
However, CRFSEG outperformed VeriFinger by reducing the MAE by 13.44, 12.21, 15.11, and 10.12 pixels in the left, right, top, and bottom sides, respectively, of detected bounding boxes on the \textit{Combined} dataset.

\begin{figure}[!ht]
    \begin{subfigure}[b]{0.33\textwidth}
    \centering
    \includegraphics[width=1\linewidth]{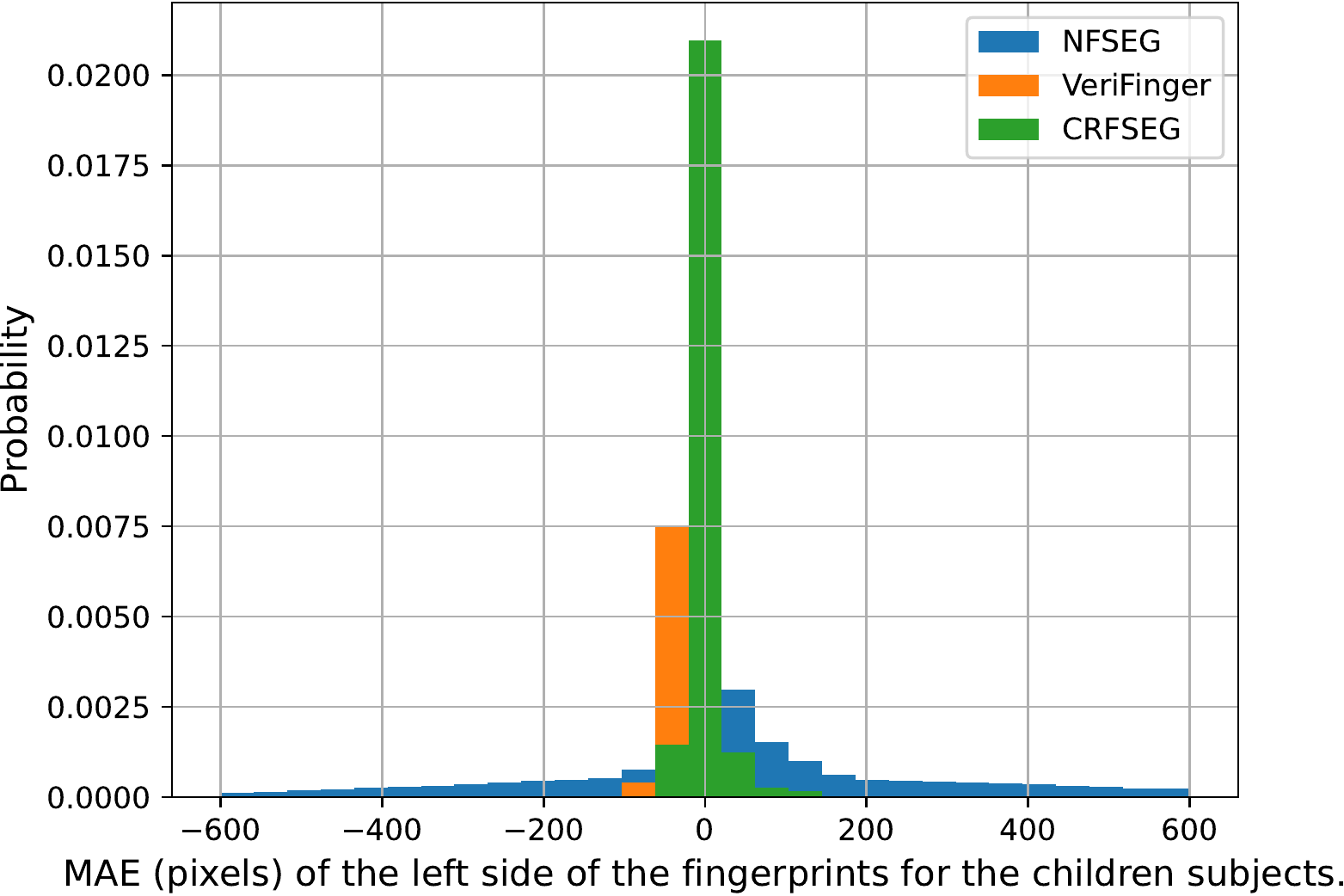}
    \label{fig:MAE_Left_Children}
    \end{subfigure}
    \begin{subfigure}[b]{0.33\textwidth}
    \centering
    \includegraphics[width=1\linewidth]{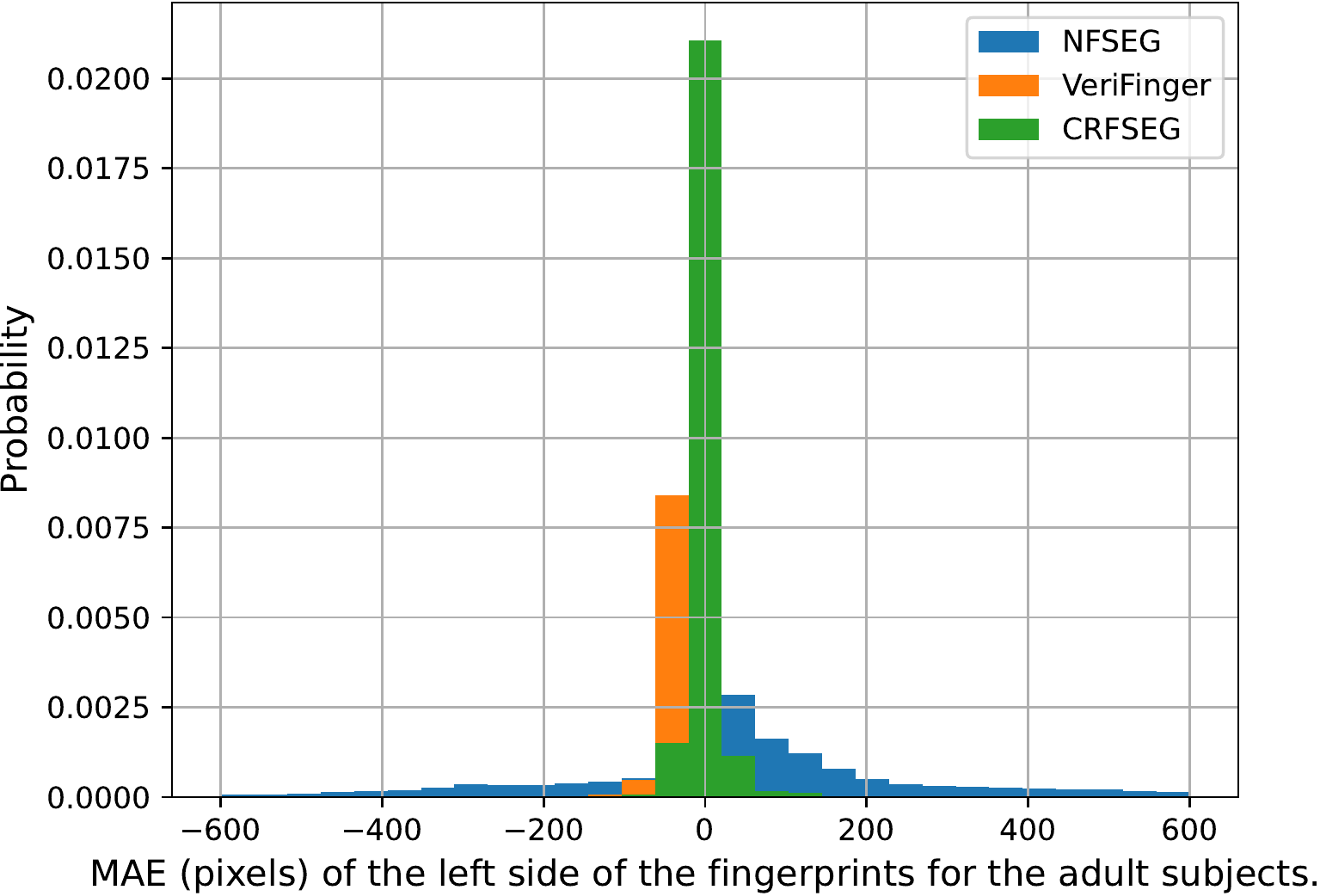}
    \label{fig:MAE_Left_adult}
    \end{subfigure}
    \begin{subfigure}[b]{0.33\textwidth}
    \centering
    \includegraphics[width=1\linewidth]{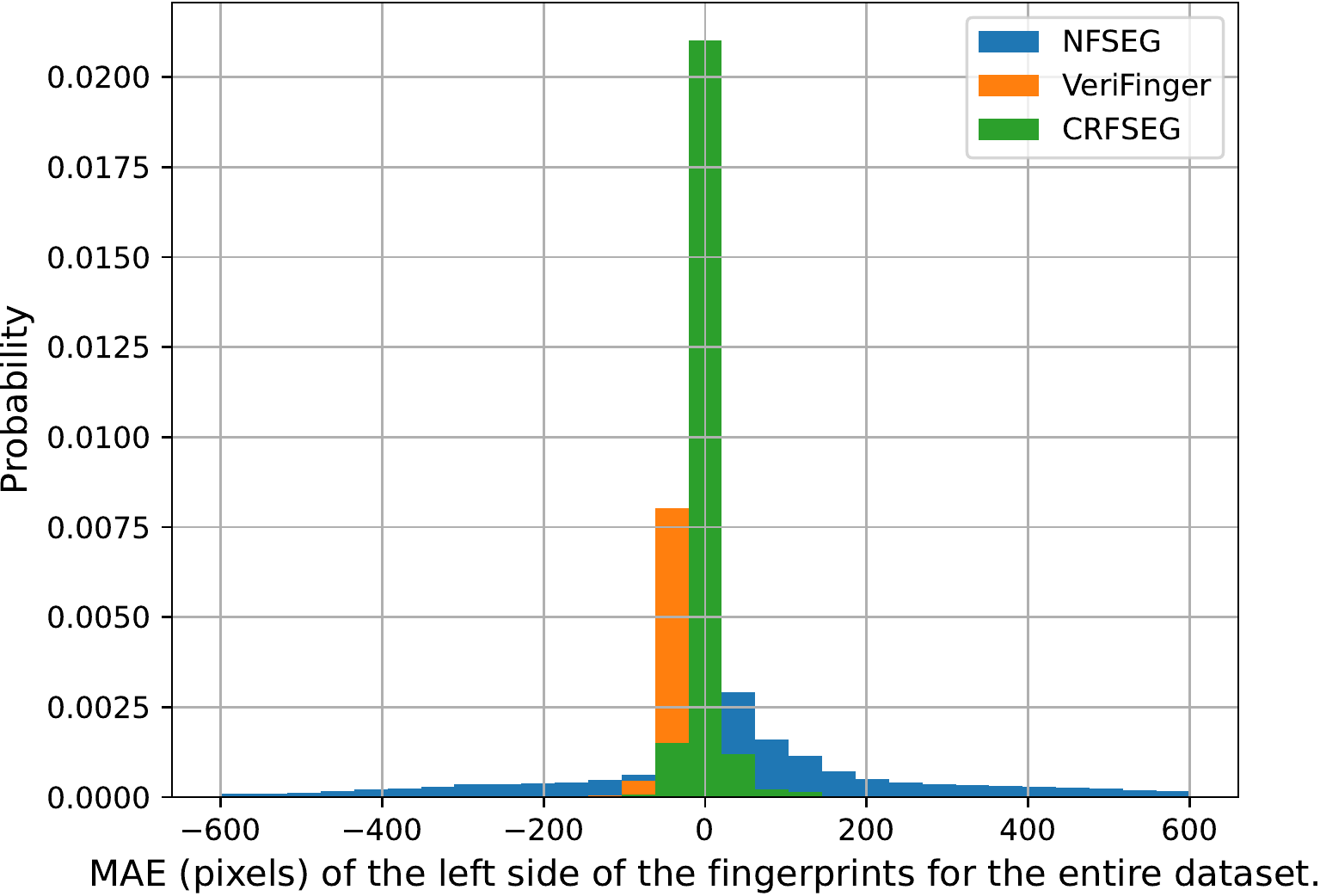}
    \label{fig:MAE_left_entire}
    \end{subfigure}
    \begin{subfigure}[b]{0.33\textwidth}
    \centering
    \includegraphics[width=1\linewidth]{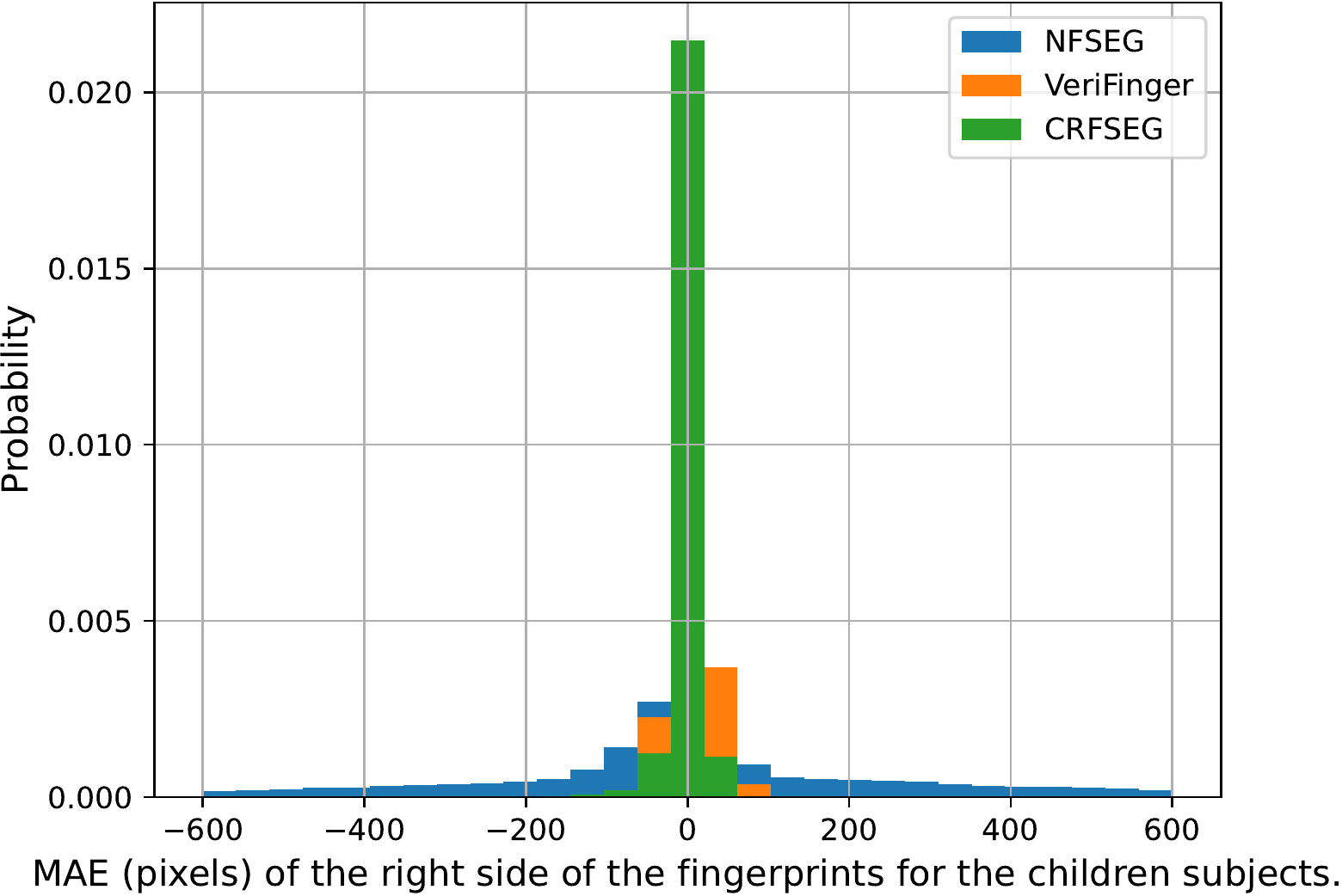}
    \label{fig:MAE_Right_Children}
    \end{subfigure}
    \begin{subfigure}[b]{0.33\textwidth}
    \centering
    \includegraphics[width=1\linewidth]{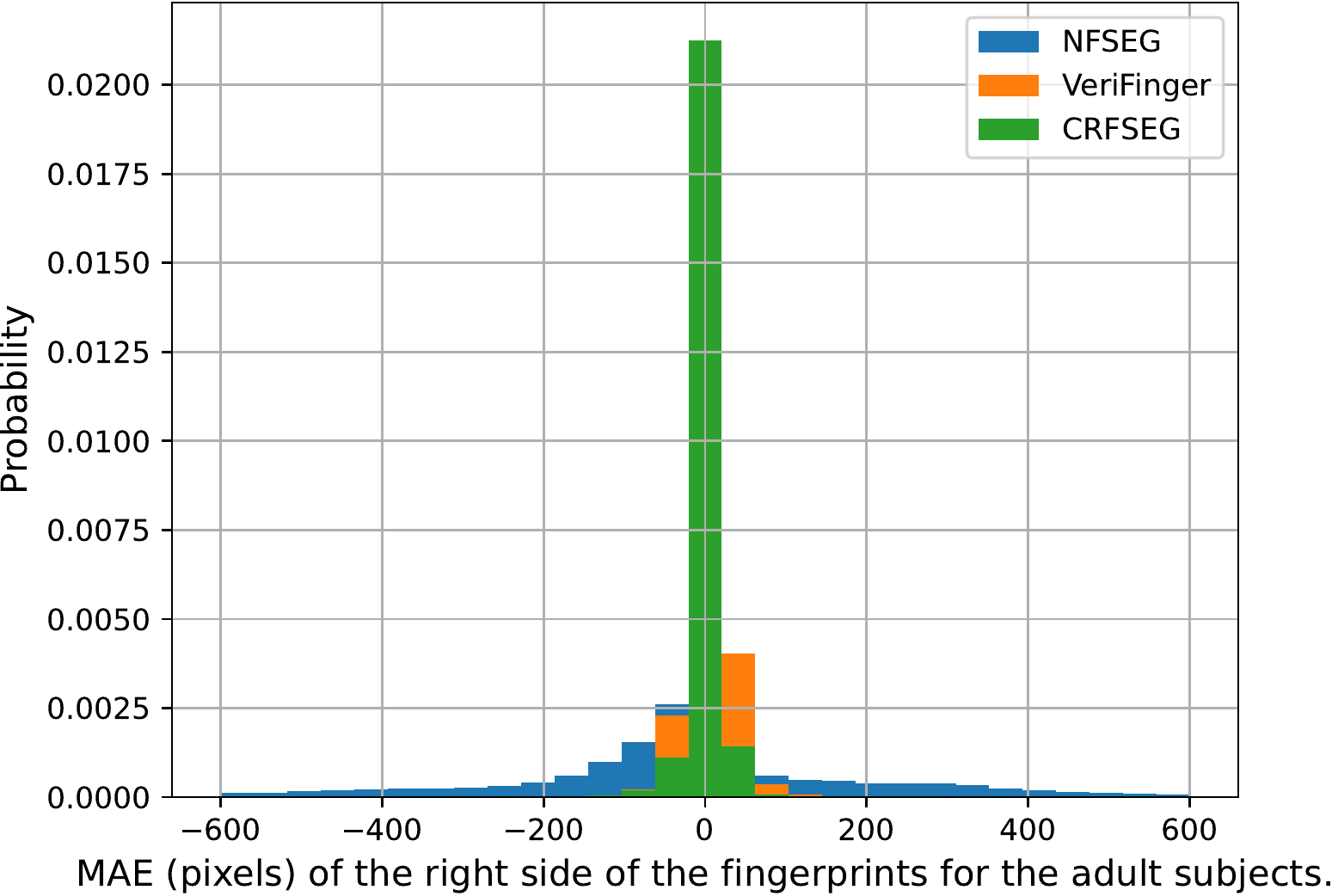}
    \label{fig:MAE_Right_adult}
    \end{subfigure}
    \begin{subfigure}[b]{0.33\textwidth}
    \centering
    \includegraphics[width=1\linewidth]{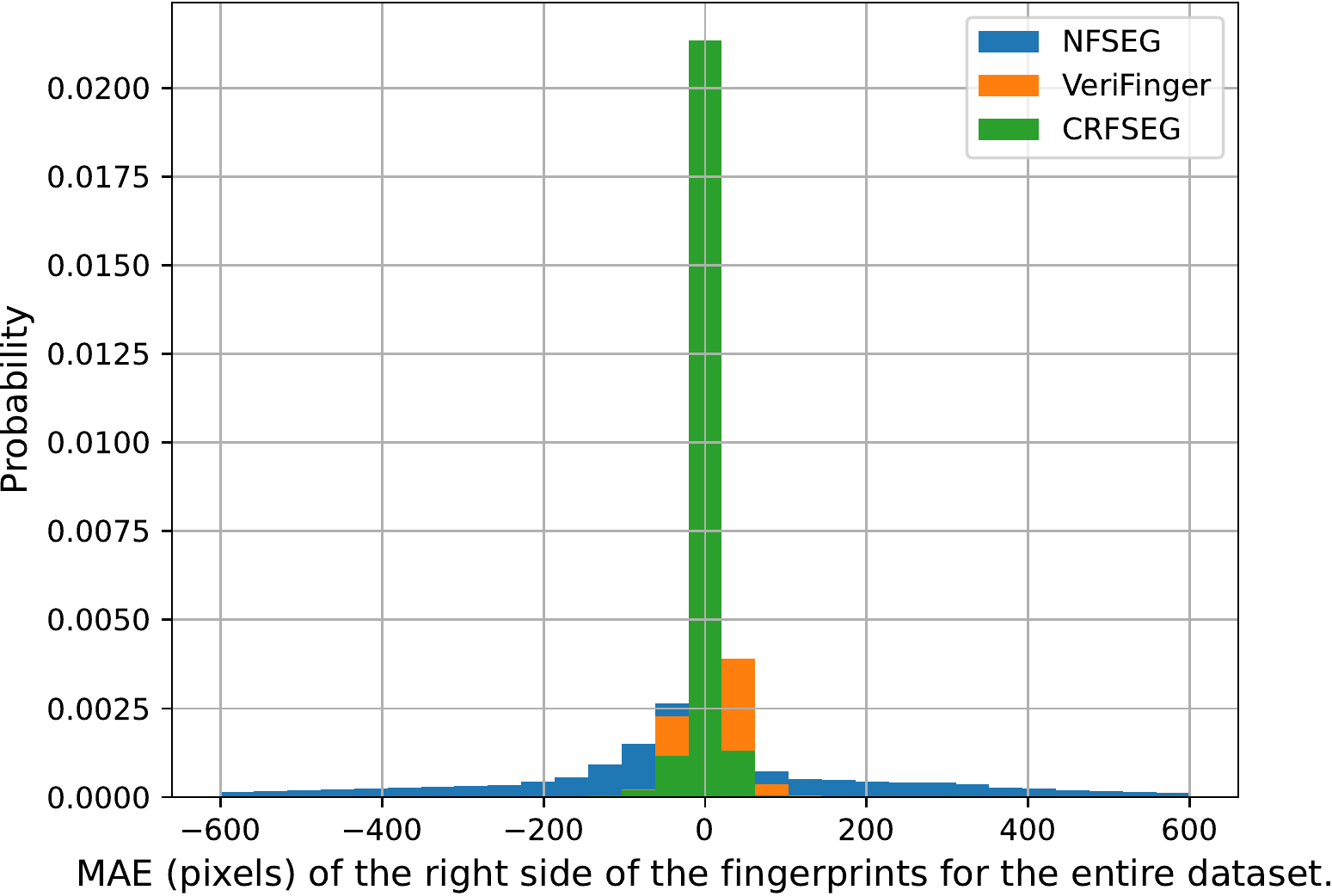}
    \label{fig:MAE_right_entire}
    \end{subfigure}
    \begin{subfigure}[b]{0.33\textwidth}
    \centering
    \includegraphics[width=1\linewidth]{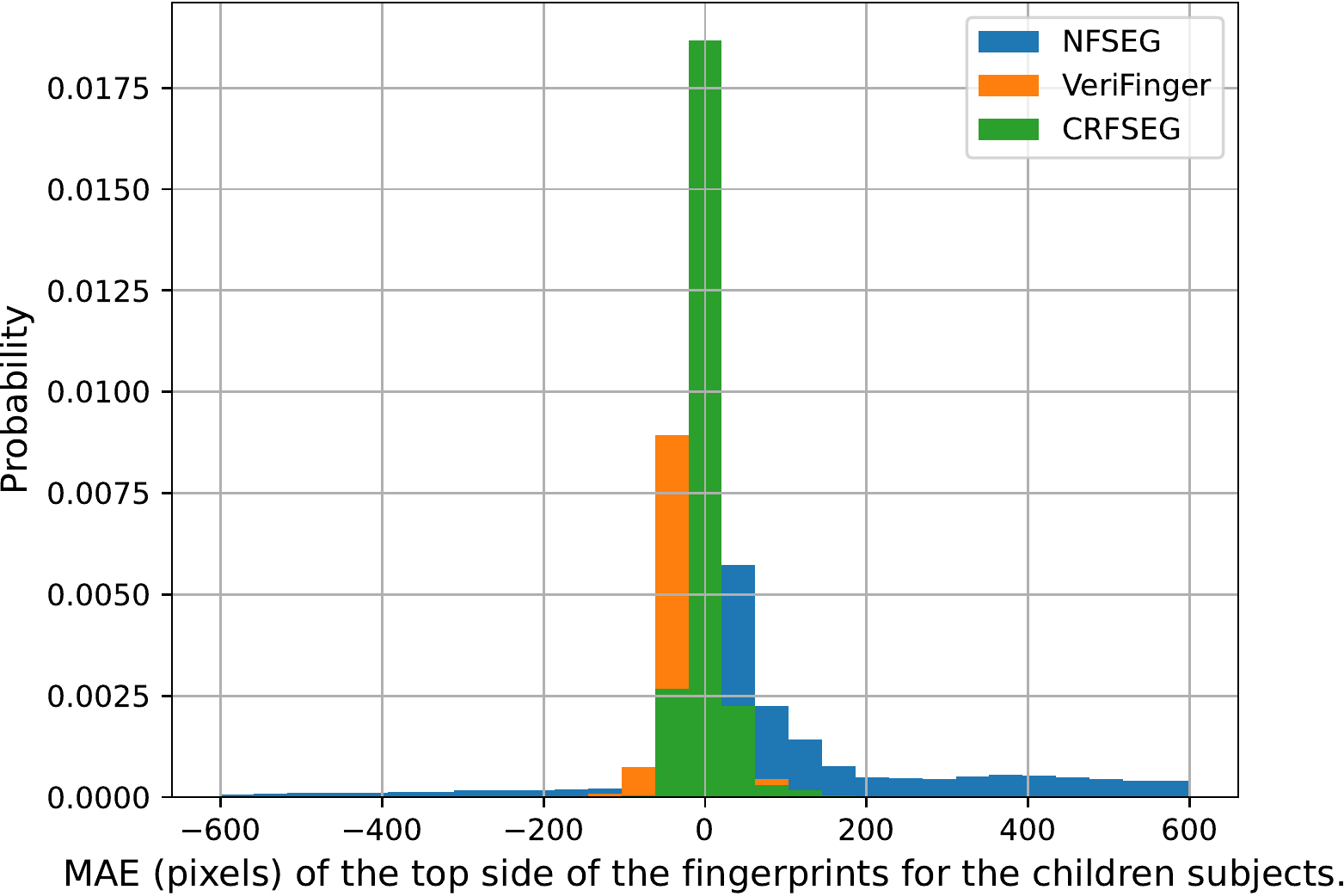}
    \label{fig:MAE_Top_Children}
    \end{subfigure}
    \begin{subfigure}[b]{0.33\textwidth}
    \centering
    \includegraphics[width=1\linewidth]{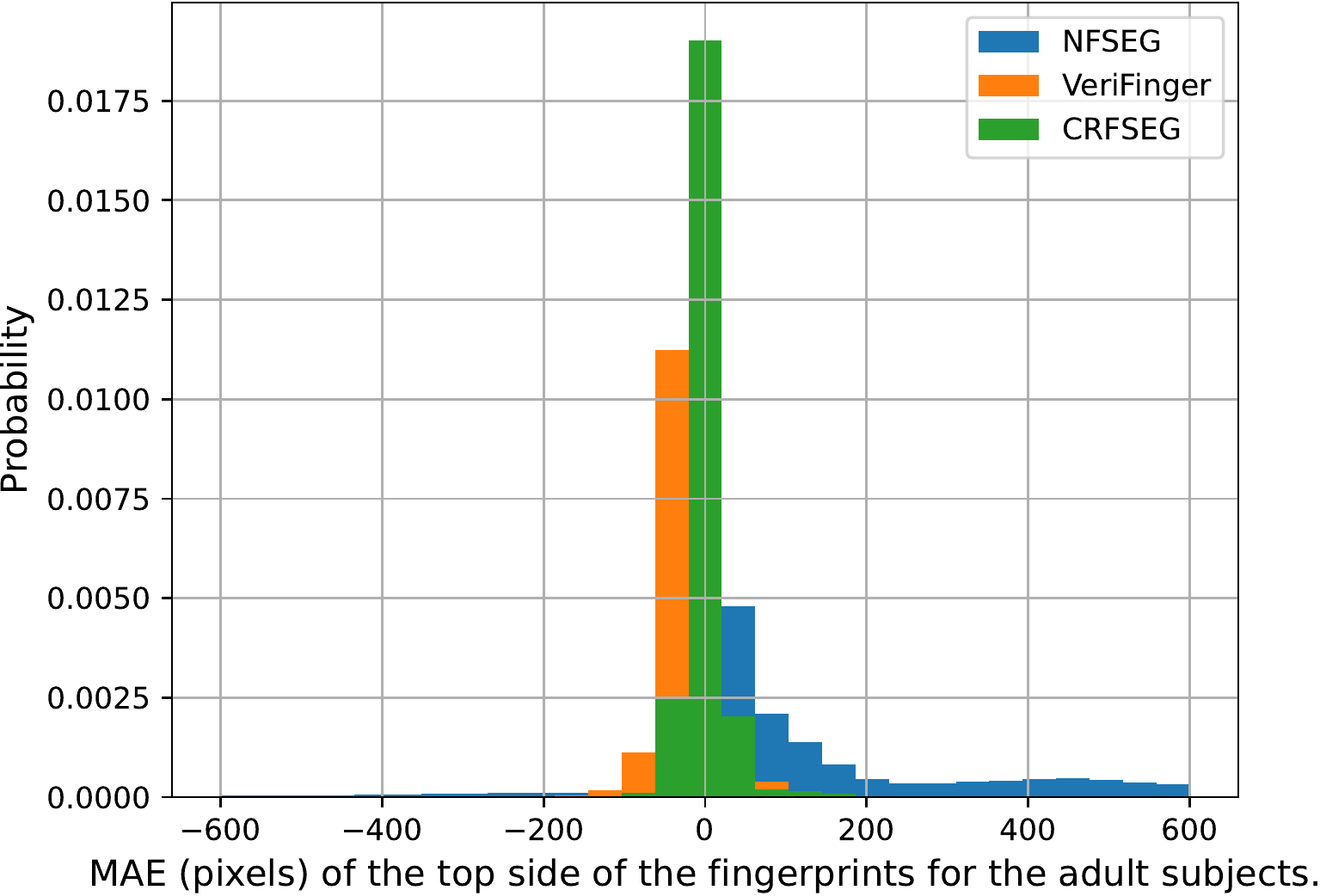}
    \label{fig:MAE_Top_adult}
    \end{subfigure}
    \begin{subfigure}[b]{0.33\textwidth}
    \centering
    \includegraphics[width=1\linewidth]{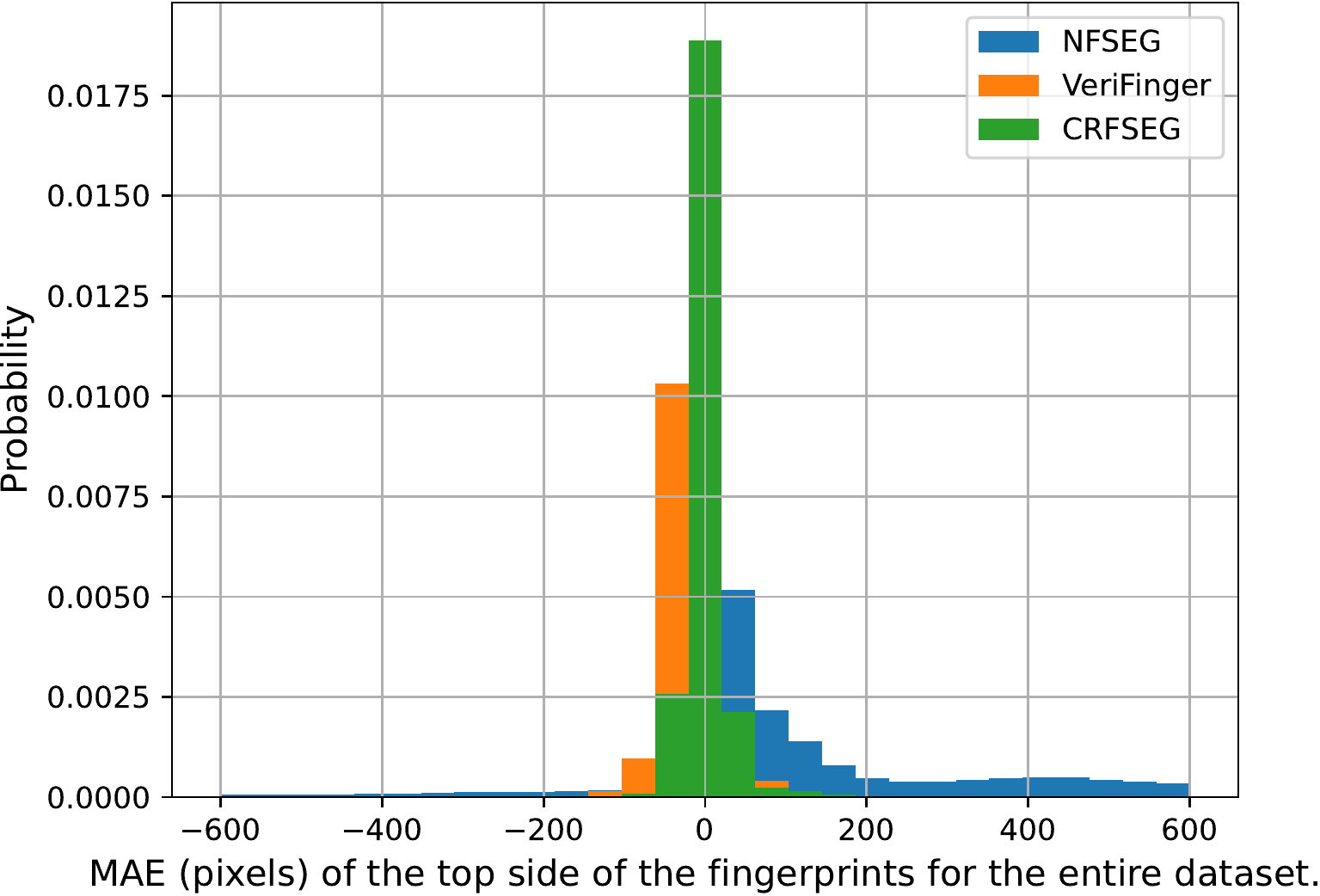}
    \label{fig:MAE_top_entire}
    \end{subfigure}
    \begin{subfigure}[b]{0.33\textwidth}
    \centering
    \includegraphics[width=1\linewidth]{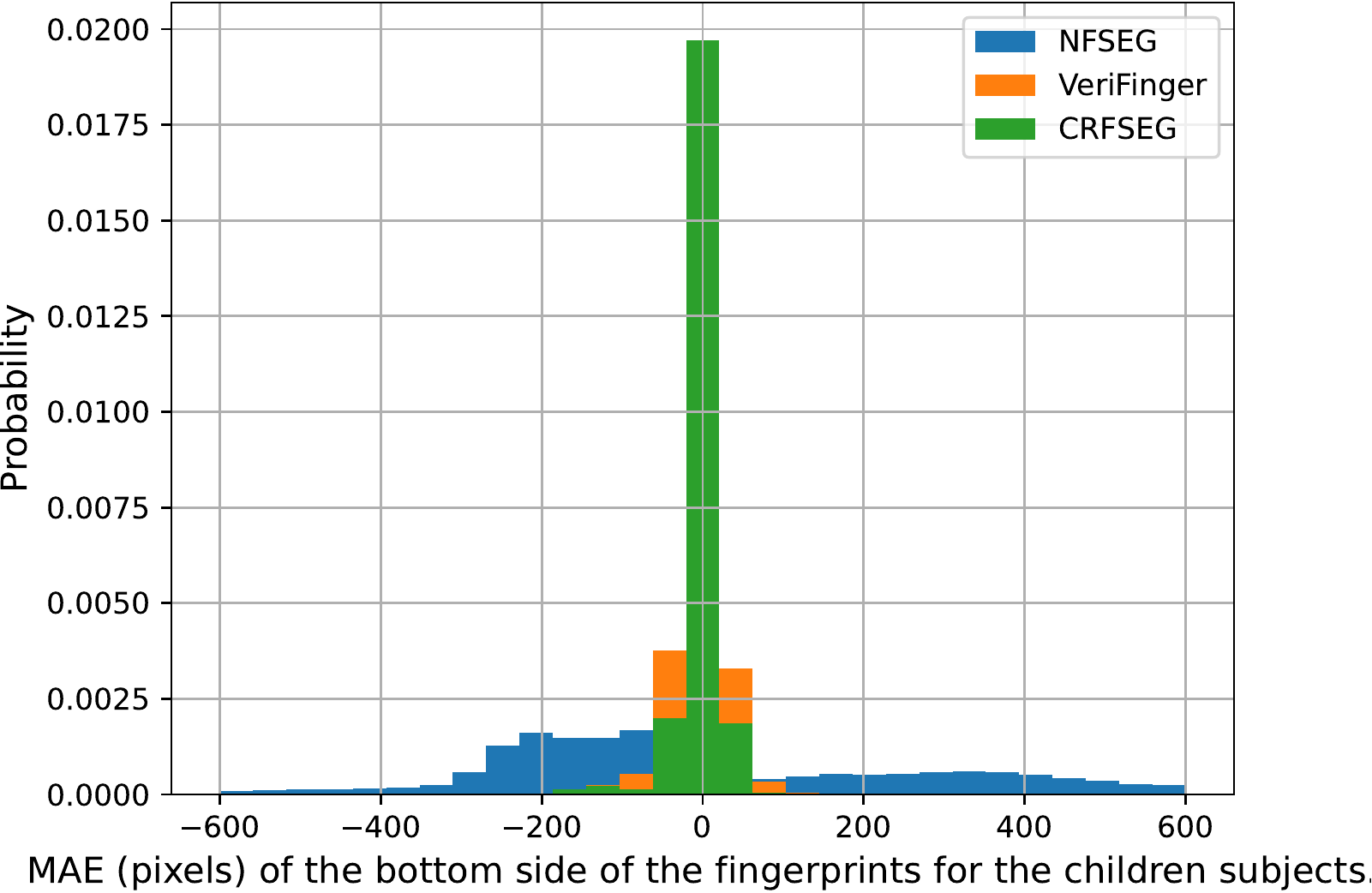}
    \label{fig:MAE_Bottom_children}
    \end{subfigure}
    \begin{subfigure}[b]{0.33\textwidth}
    \centering
    \includegraphics[width=1\linewidth]{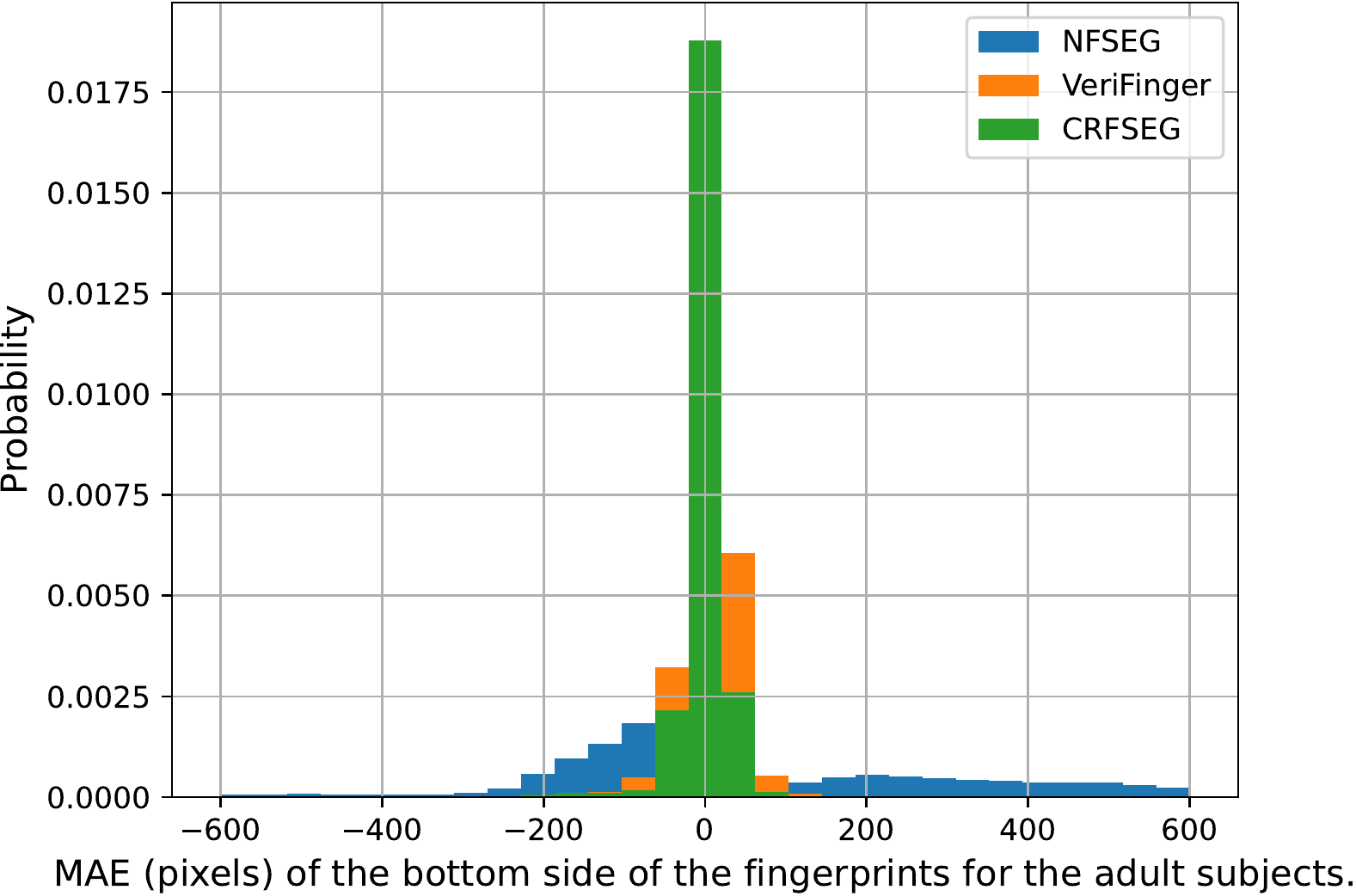}
    \label{fig:MAE_Bottom_adult}
    \end{subfigure}
    \begin{subfigure}[b]{0.33\textwidth}
    \centering
    \includegraphics[width=1\linewidth]{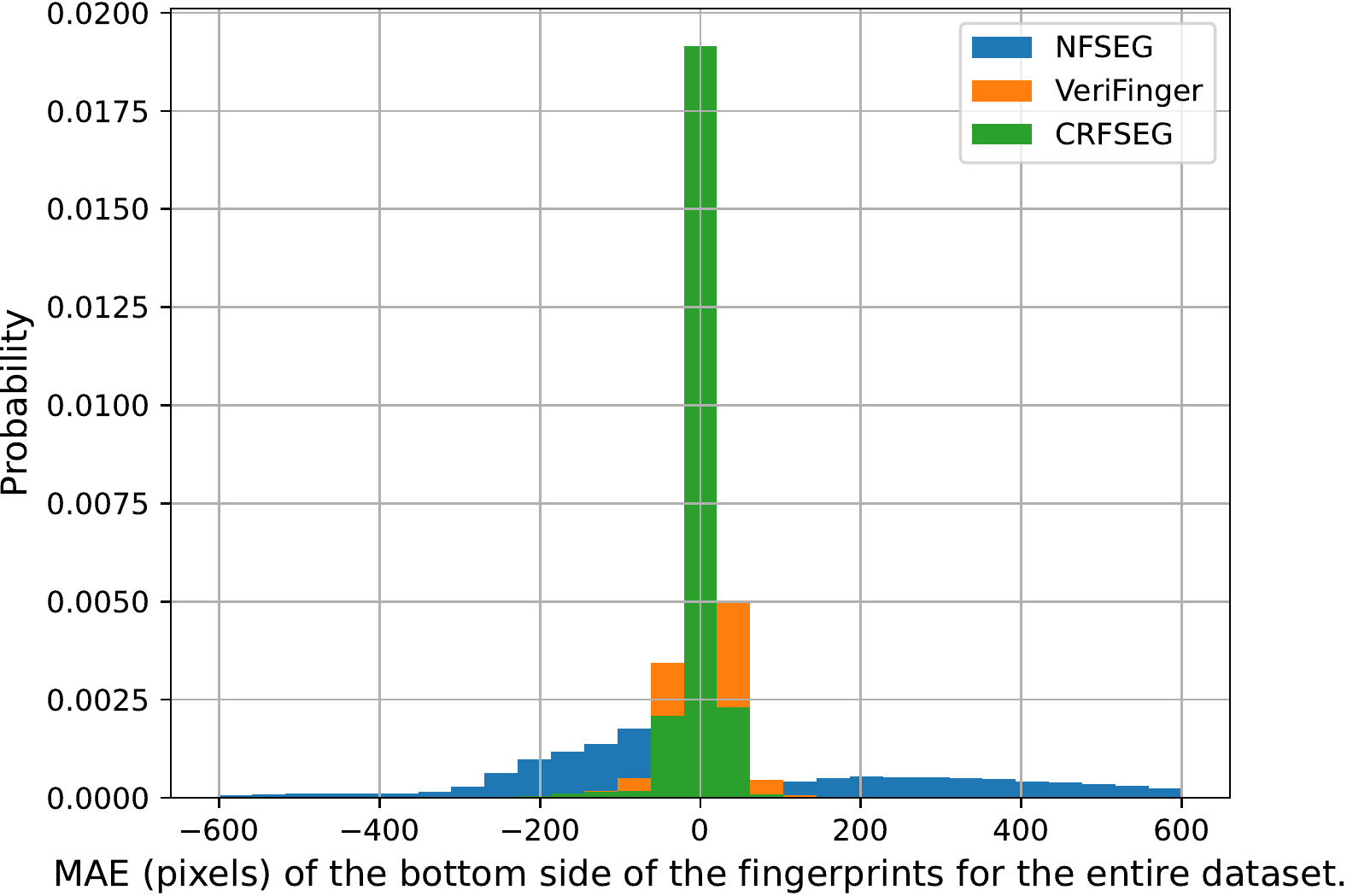}
    \label{fig:MAE_Bottom_entire}
    \end{subfigure}
  \caption{The histograms of the MAE generated from the output of different slap segmentation algorithms using \autoref{equ:MAEs}. Blue, orange, and green colors represent the MAE of NFSEG, VeriFinger, and CRFSEG, respectively, on the \textit{Combined} dataset. To evaluate segmentation algorithms on different age groups, we report MAE in children and adult groups separately. The four images in the first column depict the MAEs of each side of the bounding box generated using slap images of child subjects, the second column depicts MAEs generated using adult slaps, and the third column shows MAEs generated using the entire \textit{Combined} dataset (containing both adult and child subjects). The MAE values for the sides of the bounding boxes are presented in four rows with the first row representing the left side, the second row representing the right side, the third row representing the top side, and the fourth row representing the bottom side. The long tails of the blue histograms indicate poor performance of NFSEG. WVeriFinger performs better than NFSEG on the slaps of children and adult subjects. However, CRFSEG outperformed both NFSEG and VeriFinger in terms of MAE.}
  \label{fig:MAEFullHits}
\end{figure}

We used histograms to show, analyze and evaluate the MAE results. To generate these histograms, we used the results obtained by subtracting the coordinate position of one side of all ground-truth bounding boxes from the coordinate position of the corresponding side of the detected bounding boxes. \autoref{fig:MAEFullHits} illustrates the histograms of MAE for all sides of bounding boxes generated by three segmentation models. Blue, orange, and green histograms represent the MAE of NFSEG, VeriFinger, and CRFSEG respectively. The first and second columns of \autoref{fig:MAEFullHits} show the results when we calculated MAE using children and adult subjects separately, and the third column shows results when we calculate the MAE using our entire \textit{Combined} dataset. This separation helps us to analyze the results on children and adult subjects separately and make the model robust against age variants.

\begin{figure}[!ht]
  \centering
  \includegraphics[width=.4\linewidth]{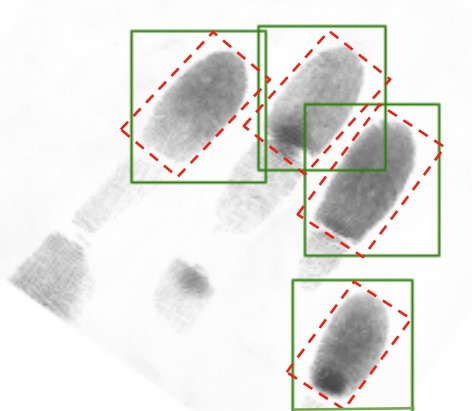}
\caption{Green rectangles show the detected bounding boxes by NFSEG and red rectangles show the ground-truth bounding box. NFSEG encountered issues with correctly rotating certain slap images, causing the bounding boxes to overlap and encompass extraneous regions beyond the intended fingerprints, which subsequently leads to lower fingerprint-matching performance.}
\label{fig:pixelLossExample}
\end{figure}

We observed that while VeriFinger and CRFSEG maintain performance on \textit{Combined} slap datasets, NFSEG is particularly susceptible to this type of MAE error, especially when slap images are over-rotated or from children subjects. 
The long blue tail of the error in histograms of the NFSEG model confirms that NFSEG over-segmented many samples beyond the Minimum Tolerance Limit (MTL) of 64 pixels suggested by Slapseg-II \cite{watson_slapsegii_2010}.
\autoref{fig:pixelLossExample} depicts an example of this problem, where the red bounding boxes indicate the ground truth and the green bounding boxes present the bounding box predicted by NFSEG.
NFSEG failed to correctly rotate some slap images causing the identified bounding boxes to encompass additional areas beyond the intended fingerprints.
This reduces the matching performance of fingerprints segmented by the NFSEG segmentation system. 
\begin{table} [ht]
    \renewcommand{\arraystretch}{1}
\caption{Mean Absolute Error (MAE) and its standard deviation for NFSEG, VeriFinger, CRFSEG on our \textit{Combined} slap dataset. It is calculated by taking the average of the absolute differences between each side of the detected bounding box and the corresponding side of the ground-truth bounding box in terms of pixels. The smaller MAE indicates better performance. }
\label{table:AvgMAE} 
  \centering
  \begin{tabular}{ccccc}
  \hline
  {} & {} & {} & {Model} & {} \\
   \cline{3-5}
{} & {} & {NFSEG} & {VeriFinger} & {CRFSEG} \\
\cline{3-5}
Dataset & Side & MAE (Std.) & MAE (Std.) & MAE (Std.) \\
\hline 
\multirow {4}{*}{Children} & {Left} & {184.74} (274.35) & {28.63} (30.53) & \textbf{17.58 (20.55)} \\

 \cline{2-5}
 {} & {Right} & {182.08} (272.16) & {27.27} (31.34)  & \textbf{16.54 (17.31)} \\
 \cline{2-5}
{} & {Top} & {240.22} (344.53) & {30.16} (35.60) & \textbf{18.06 (24.89)} \\
 \cline{2-5}
{}  &  {Bottom} & {260.73} (382.92)  & {25.87} (35.64) & \textbf{17.19 (26.56)} \\
 \hline

\multirow {4}{*}{Adult} & {Left} & {145.29} (215.80)  &  {34.11} (26.81) & \textbf{19.08 (20.20)} \\
 \cline{2-5}
{}  & {Right} & {144.13} (215.97) & {31.79} (27.11) & \textbf{18.58 (18.56)} \\
  \cline{2-5}
 {} & {Top} & {193.61} (301.27) & {35.59} (36.74) & \textbf{18.47 (26.36)} \\
 \cline{2-5}
{}  &  {Bottom} & {184.24} (306.34)  & {30.05} (36.73) & \textbf{18.96 (28.24)} \\
\hline

\multirow {4}{*}{Entire} & {Left} & {161.13} (241.05) & {31.92} (28.37) & \textbf{18.48 (20.35)} \\
 \cline{2-5}
{}  & {Right} & {159.36} (240.18) & {29.98} (28.88) & \textbf{17.77 (18.08)} \\
  \cline{2-5}
{}  & {Top} & {212.32} (319.78) & {33.42} (36.36) & \textbf{18.31 (25.78)} \\
  \cline{2-5}
{}  & {Bottom} & {214.95} (339.21) & {28.37} (36.50) & \textbf{18.25 (27.59)} \\
\hline
\end{tabular}
\end{table}

\subsubsection*{Error in angle prediction (EAP)}
The EAP is the difference between the angle predicted by the segmentation algorithms and the ground-truth angle of a slap image. EAP is directly related to the preciseness of bounding boxes detected by the algorithms. Less deviation between predicated angles and ground-truth angles produces more accurate segmentation results. \autoref{equ:EAP} is used to calculate EAP. \autoref{table:AvgAngleError} shows the EAP and its standard deviation which is calculated by using the predicted angle output of different segmentation algorithms on the \textit{Combined} dataset. The EAP values of CRFSEG are lower than the EAP values of NFSEG and VeriFinger, indicating less error during angle prediction, resulting in better performance. 
\begin{table} [ht]
    \renewcommand{\arraystretch}{1}
\caption{The Error in Angle Prediction (EAP) is obtained by finding the difference between the angle predicted by the segmentation algorithms and the ground-truth angle of a slap image. This table presents the average EAP and its standard deviation for three algorithms, NFSEG, VeriFinger, and CRFSEG, calculated on the \textit{Combined} slap dataset. The smaller EAP and lower standard deviation are the indicators of  better segmentation results. }
\label{table:AvgAngleError} 
  \centering
  \begin{tabular}{cccc}
    \hline 
{} & {} & {Model} & {} \\
   \cline{2-4}
{} & {NFSEG} & {VeriFinger} & {CRFSEG} \\
\cline{2-4}
 Dataset & EAP (Std.) & EAP (Std.) & EAP (Std.)\\
\hline 
Children & {$33.18^\circ$} (47.85) & {$9.88^\circ$} (14.14) & {$\textbf{6.66}^\circ$} (9.81) \\
\cline{1-4}
Adult & {$29.3^\circ$} (43.83)  & {$7.65^\circ$} (11.13) & {$\textbf{5.27}^\circ$} (7.34) \\
\cline{1-4}
Entire & {$31.62^\circ$} (46.56)  & {$8.61^\circ$} (12.23) & {$\textbf{5.86}^\circ$} (8.05) \\
\hline
\end{tabular}
\end{table}

\begin{figure}[ht]
  \begin{subfigure}[b]{0.33\textwidth}
    \centering
    \includegraphics[width=1\linewidth]{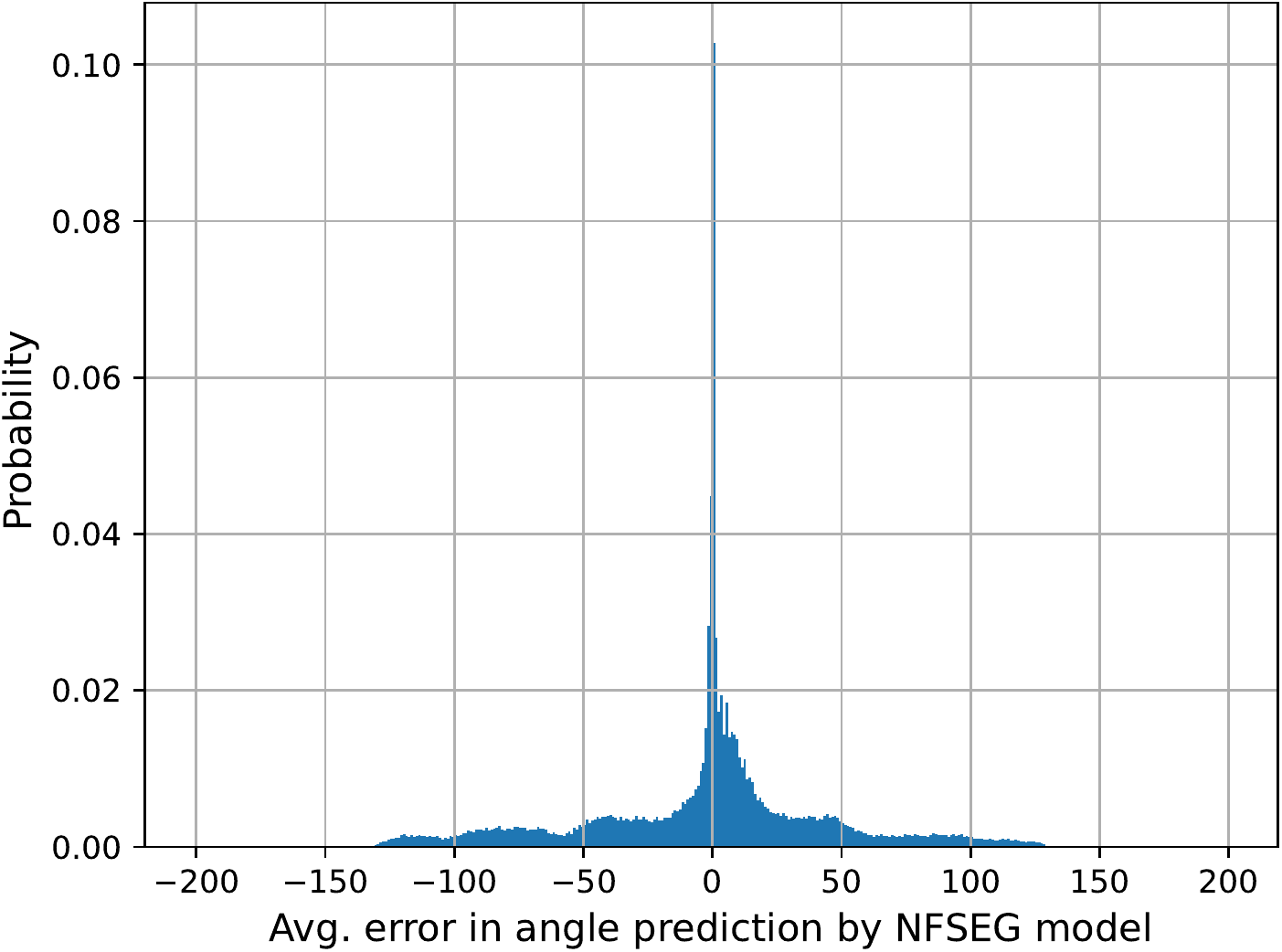}
    \label{fig:NFSEGAvgAngleErrorhis}
    \end{subfigure}
    \begin{subfigure}[b]{0.33\textwidth}
    \centering
    \includegraphics[width=1\linewidth]{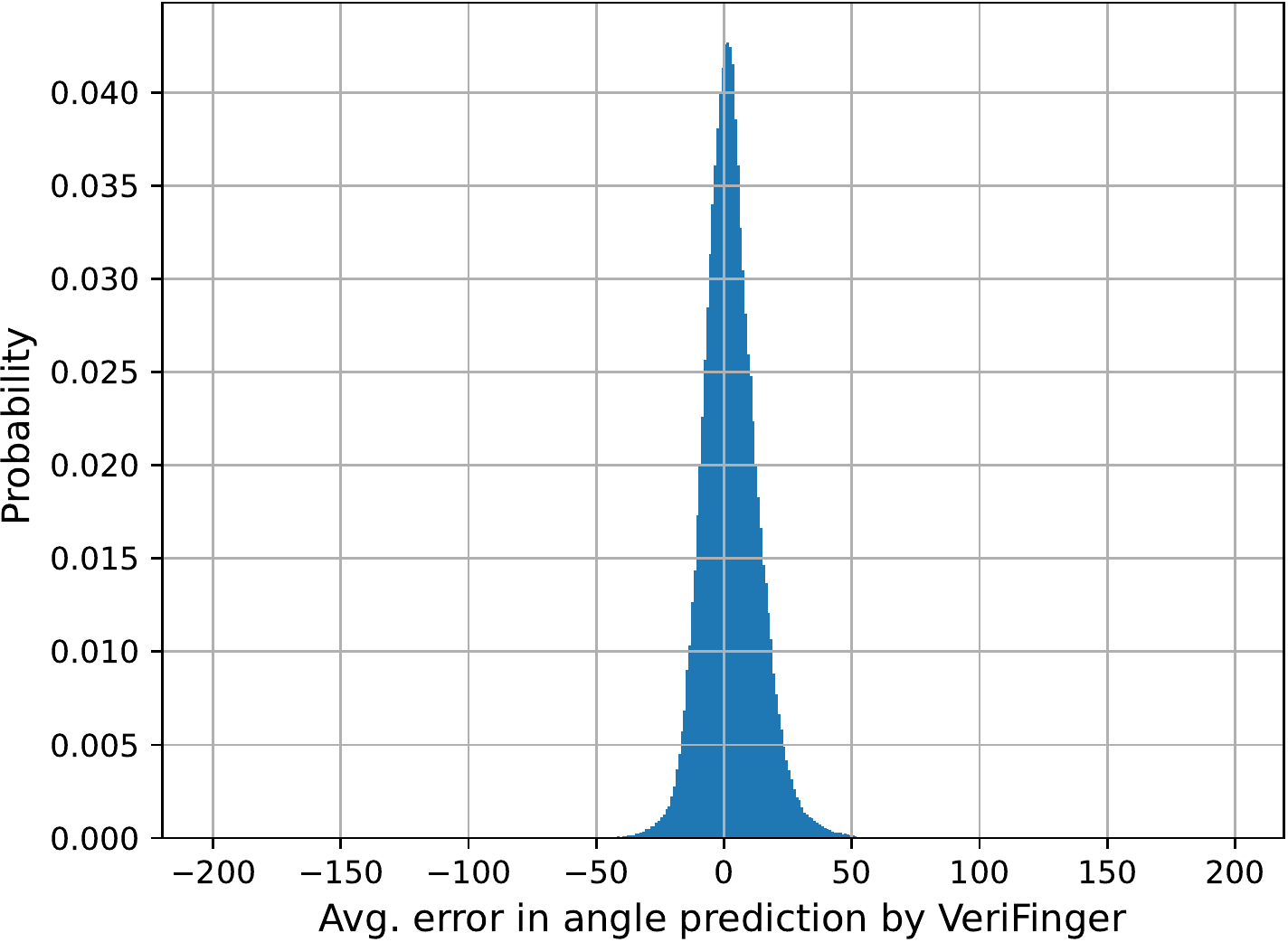}
    \label{fig:VeriAvgAngleErrorhis}
    \end{subfigure}
    \begin{subfigure}[b]{0.33\textwidth}
    \centering
    \includegraphics[width=1\linewidth]{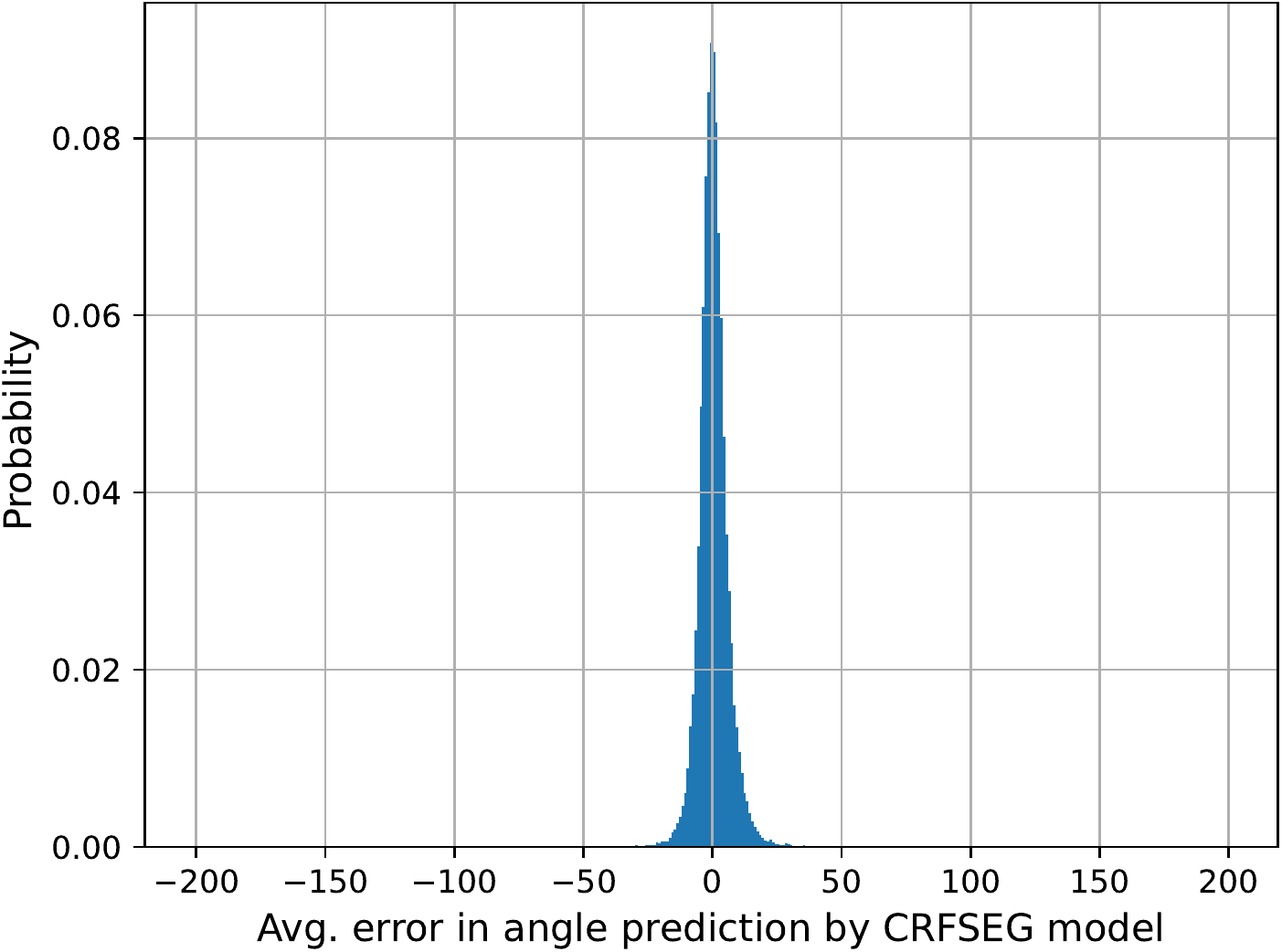}
    \label{fig:CRFSEGAvgAngleErrorhis}
    \end{subfigure}
  \caption{The histograms of the error in slap angle prediction by different segmentation models on the \textit{Combined} dataset. This error is calculated by subtracting the angles predicted by the models from the ground-truth angles of the slap images. Low standard deviation values indicate better results.}
  \label{fig:AvgAngleErrorhis}
\end{figure}
\begin{figure}
\centering
\begin{subfigure}[b]{0.33\textwidth}
\centering
\includegraphics[width=1\linewidth]{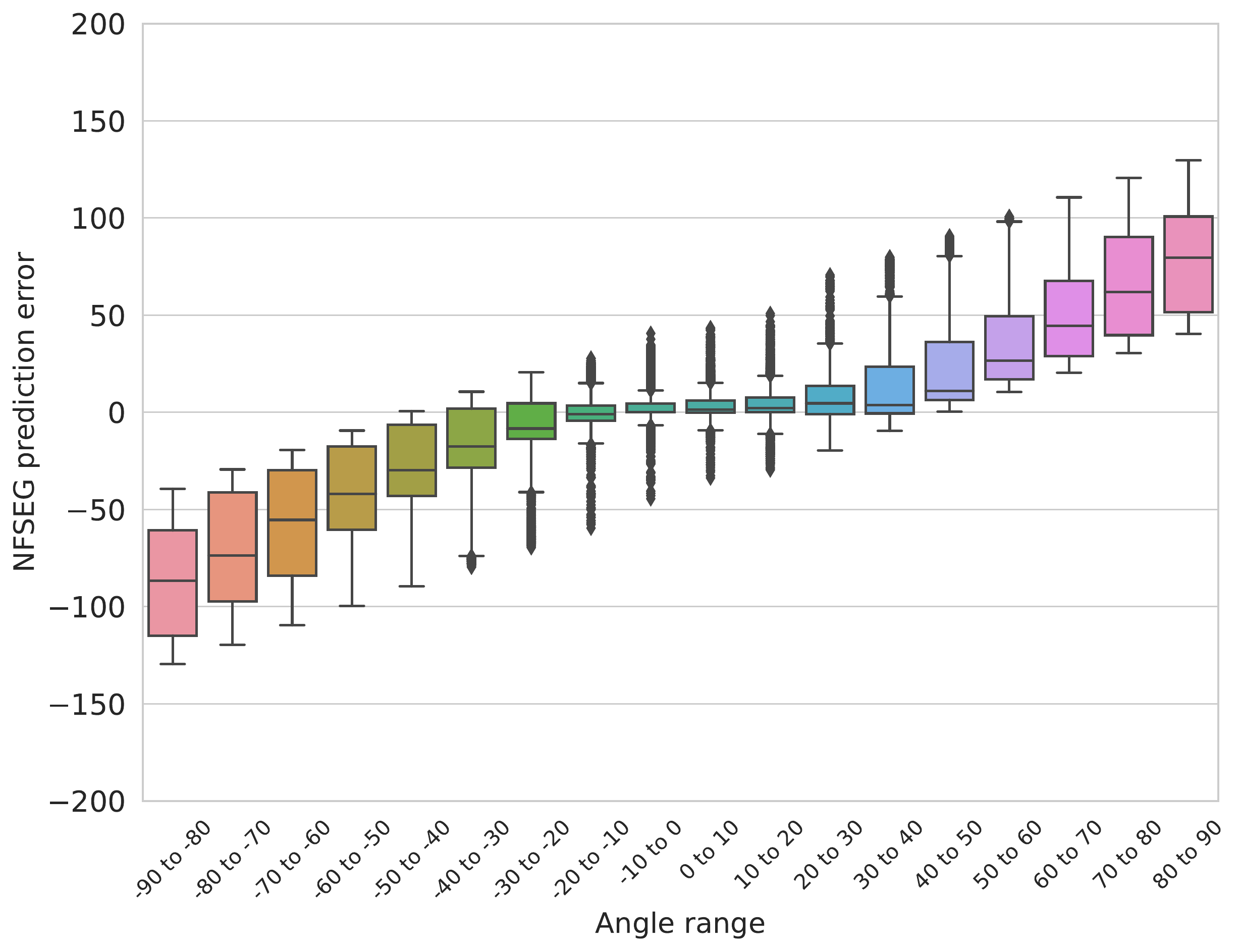}
\caption{EAP of NFSEG.}
\label{fig:nfseg_boxplot_angle_errors}
\end{subfigure}
\hfill
\begin{subfigure}[b]{0.33\textwidth}
\centering
\includegraphics[width=1\linewidth]{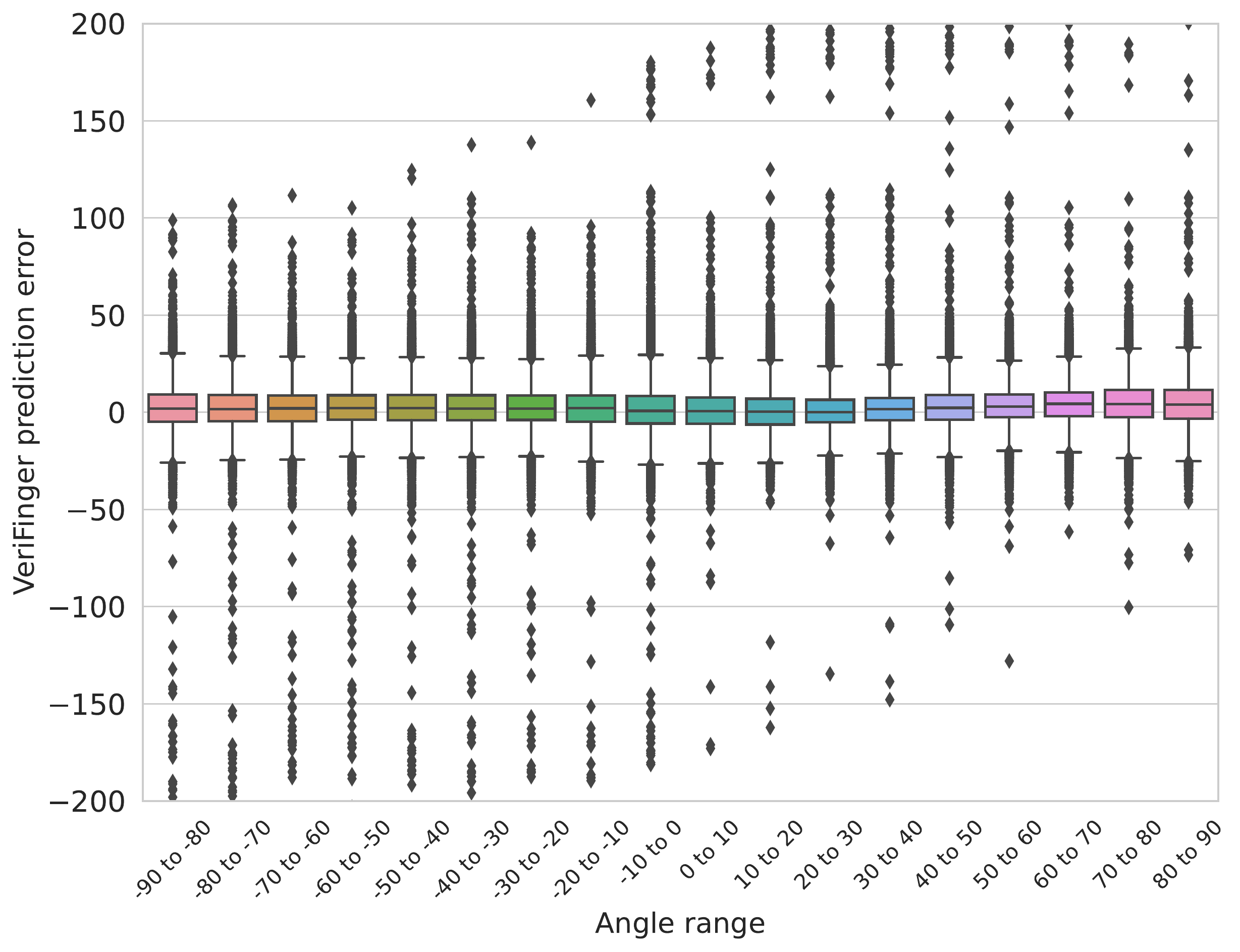}
\caption{EAP of VeriFinger.}
\label{fig:veri_boxplot_angle_errors}
\end{subfigure}
\hfill
\begin{subfigure}[b]{0.33\textwidth}
\centering
\includegraphics[width=1\linewidth]{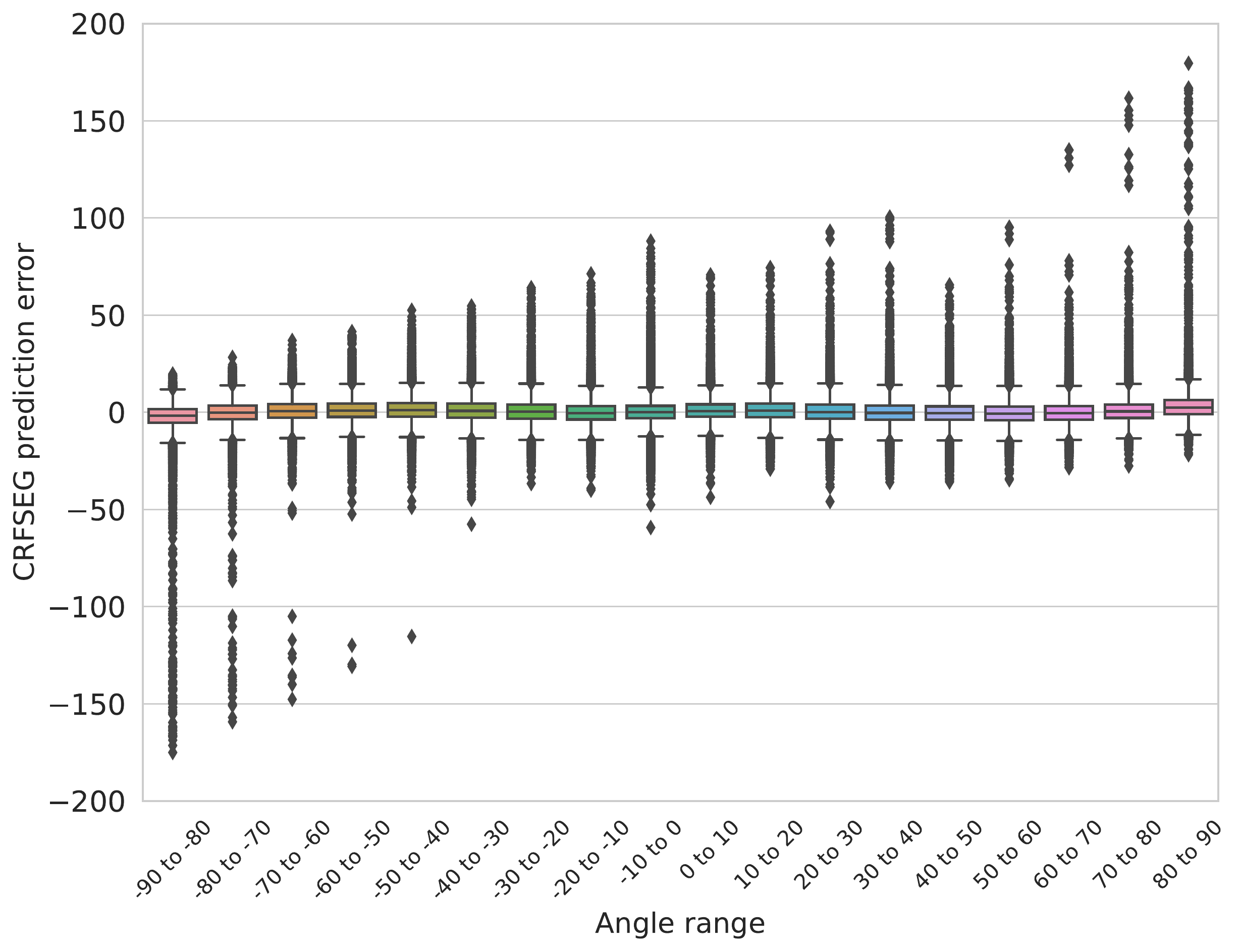}
\caption{EAP of CRFSEG.}
\label{fig:crfseg_boxplot_angle_errors}
\end{subfigure}
 \caption{Boxplots of the error in angle prediction (EAP) of different segmentation algorithms on the \textit{Combined} dataset. The boxplot statistical analysis of the mean with
 $\pm 10^\circ$ (standard errors) for the EAP values of VeriFinger and CRFSEG indicates that these algorithms are invariant to slap rotation. On the contrary, the mean of NFSEG output is high and not consistent with the rotation of the slap images which indicates the low performance of NFSEG when slap images are overly rotated.}
 \label{fig:boxplot_angle_error}
\end{figure}

\autoref{fig:AvgAngleErrorhis} shows the histograms of the error in angle prediction of three segmentation algorithms. The standard deviation and the area of the CRFSEG histogram is significantly smaller than the NFSEG and VeriFinger histograms, which indicates better performance of the CRFSEG model during angle prediction.
In \autoref{fig:boxplot_angle_error}, the boxplot graphs are used to visually display the minimum, the lower quartile, the median, the upper quartile, and the maximum quartile of the EAP values produced by NFSEG, VeriFinger, and CRFSEG. The line that is drawn across the box represents the median value of EAP. The boxplot graph shows that the absolute median value of EAP produced by the CRFSEG and VeriFinger models is lower than the absolute median value of EAP produced by NFSEG. Hence, it is experimentally
shown that the NFSEG gives lower accuracy in estimating the slap angles as compared to the VeriFinger and CRFSEG. 

\subsubsection*{Label prediction accuracy}
In order to calculate the fingerprint label prediction accuracy of NFSEG, VeriFinger, and CRFSEG models, we use the 1-Hamming Loss metric, a common measure for multi-label classifiers \cite{Khamis15WTmultilebel}.
We individually study and analyze how the prediction performance varies with age group by separating children and adult subjects on our \textit{Combined} dataset.
\autoref{table:LabelPrediction} shows the overall label prediction accuracy of NFSEG, VeriFinger, and CRFSEG.

CRFSEG achieves a label accuracy of  $98.90\%$ on our \textit{Combined} dataset which is slightly lower than the accuracy (99.11\%) of NFSEG. However, it is important to mention that, the NFSEG and VeriFinger need hand information such as left hand, right hand, or thumb to segment a slap image. Without this hand information, the accuracy of NFSEG is very poor (around 15\%). On the contrary, the CRFSEG model segment slap images without any hand information. This characteristic of CRFSEG makes this model more suitable to use in situations where additional information such as hand information is not available. It is worth mentioning that someone can use NFSEG and VeriFinger for labeling slap images if the hand information is available.   
\begin{table} [ht]
    \renewcommand{\arraystretch}{1}
\caption{Label prediction accuracy of NFSEG, VeriFinger, and CRFSEG on the \textit{Combined} dataset. We also calculate label prediction accuracy separately for adult and child subjects which helps us evaluate the performance of different slap segmentation models across different age groups. Although NFSEG had slightly better label accuracy than CRFSEG, but this was due to the use of hand information (right, left, or thumb) during segmentation, without which its performance drops significantly to around 15\%. VeriFinger also requires hand information to segment a slap image, while CRFSEG can perform segmentation without the need for any additional hand information.}
\label{table:LabelPrediction} 
  \centering
  \begin{tabular}{|c|c|c|c|}
    \hline 
     {Model} & {Input} & {Dataset} & {Accuracy (\%)} \\
    \hline
    \multirow {3}{*}{\centering NFSEG} & \multirow {3}{*}{\centering Slap \& Hand Information} & {Children} & {98.26} \\
    \cline{3-4}
    {} & {} & {Adult} & {\textbf{99.95}} \\
    \cline{3-4}
    {} & {} & {Entire} & {\textbf{99.11}} \\
    \hline
    \multirow {3}{*}{\centering VeriFinger} & \multirow {3}{*}{\centering Slap \& Hand Information} & {Children} & {90.34} \\
    \cline{3-4}
    {} & {} & {Adult} & {91.03} \\
    \cline{3-4}
    {} & {} & {Entire} & {90.69} \\
    \hline
    \multirow {3}{*}{\centering CRFSEG} & \multirow {3}{*}{\centering Only Slap} & {Children} & {99.57} \\
    \cline{3-4}
    {} & {} & {Adult} & {98.48} \\
    \cline{3-4}
    {} & {} & {Entire} & {98.90} \\
    \hline
    \end{tabular}
\end{table}

Furthermore, we calculate the label prediction accuracy for the adult and children's fingerprints separately. Our results show that the label prediction accuracy of CRFSEG is consistently good for adult and children subjects. We also observe that the label prediction accuracy of NFSEG and VeriFinger is also good on adult and children slap images. However, this is because we provide hand information to NFSEG and VeriFinger during label-predicting experiments. 

\subsubsection*{Fingerprint matching}
The main goal of a fingerprint segmentation algorithm is to improve the matching performance. In our experiments, we used the VeriFinger fingerprint matcher (version 10, compliant with NIST MINEX \cite{watson2014fingerprint}) to measure the matching accuracy and quantitatively evaluate the contributions of the segmentation algorithms on the overall matching performance. We then compared the matching accuracy of CRFSEG with the state-of-the-art NFSEG and VeriFinger segmentation software. 

In order to calculate the matching accuracy, four sets of segmented fingerprint images are generated from slap images of the \textit{Combined} dataset. 
The bounding box data generated by human annotators (ground truth), NFSEG, Verifigner, and CRFSEG are used to generate those sets of segment fingerprint images. 
For every fingerprint in a segmented fingerprint set, we evaluated all the mated comparisons for our genuine distribution (3,188,232), while randomly selecting 20 non-mated fingerprints to construct an imposter distribution (32,281,840 imposter comparisons).
A total of 35,470,072 comparisons using all ten fingers are performed in our experiment to estimate the matching accuracy. To evaluate the performance of different segmentation models on child and adult fingerprints, we calculated fingerprint-matching accuracy on child and adult subjects separately. 

\begin{figure}[!ht]
  \centering
  \includegraphics[width=0.9\linewidth]{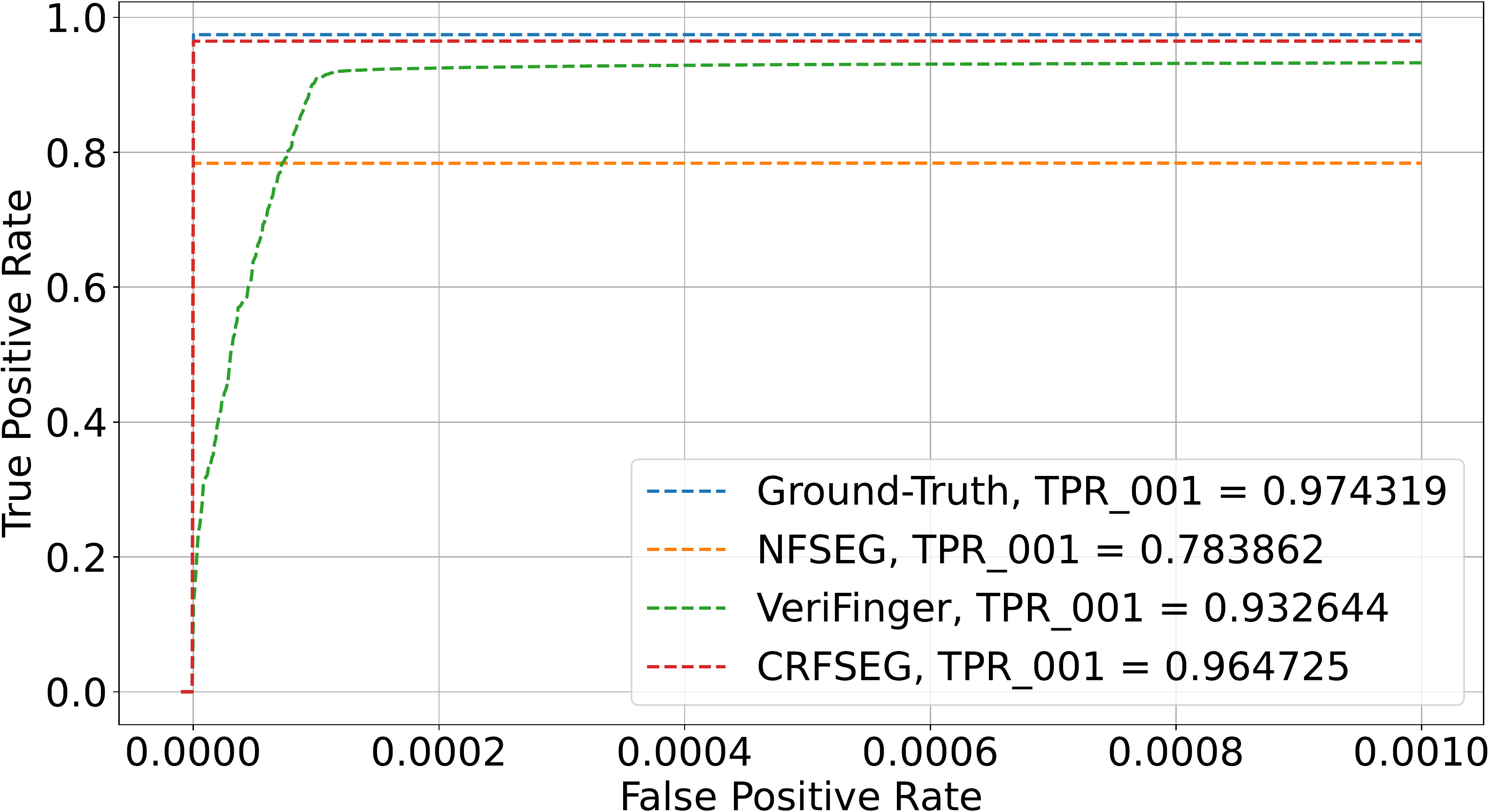}
\caption{
The Receiver Operating Characteristics (ROC) for the fingerprint matching performance of Ground Truth, NFSEG, VeriFinger, and CRFSEG on children subjects in the \textit{Combined} dataset.}
\label{fig:matching_roc_children}
\end{figure}

\begin{figure}[!ht]
  \centering
  \includegraphics[width=0.9\linewidth]{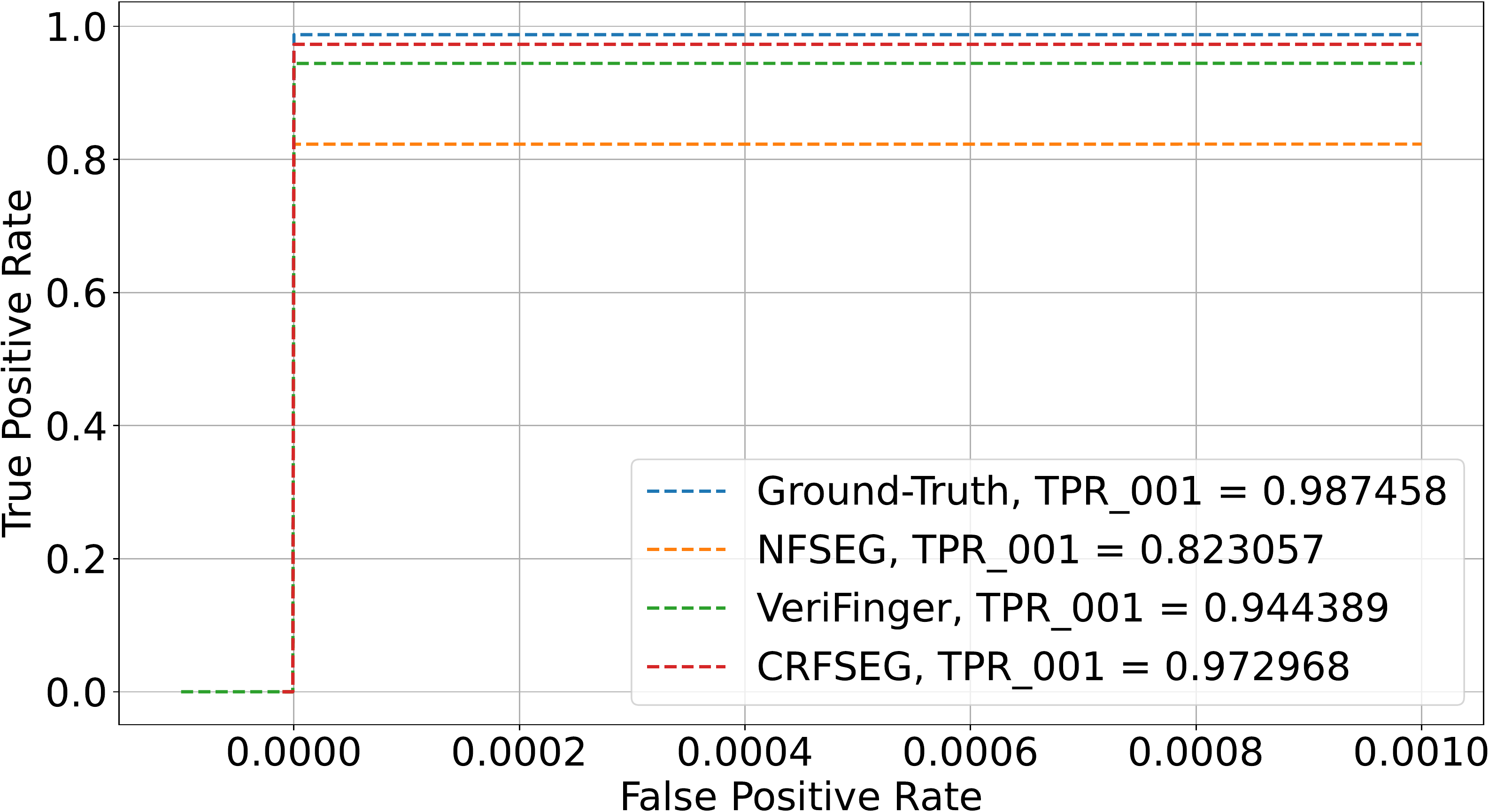}
\caption{The Receiver Operating Characteristics (ROC) for Ground Truth, NFSEG, VeriFinger, CRFSEG on adult subjects in the \textit{Combined} dataset.}
\label{fig:matching_roc_adult}
\end{figure}

\begin{figure}[!ht]
  \centering
  \includegraphics[width=0.9\linewidth]{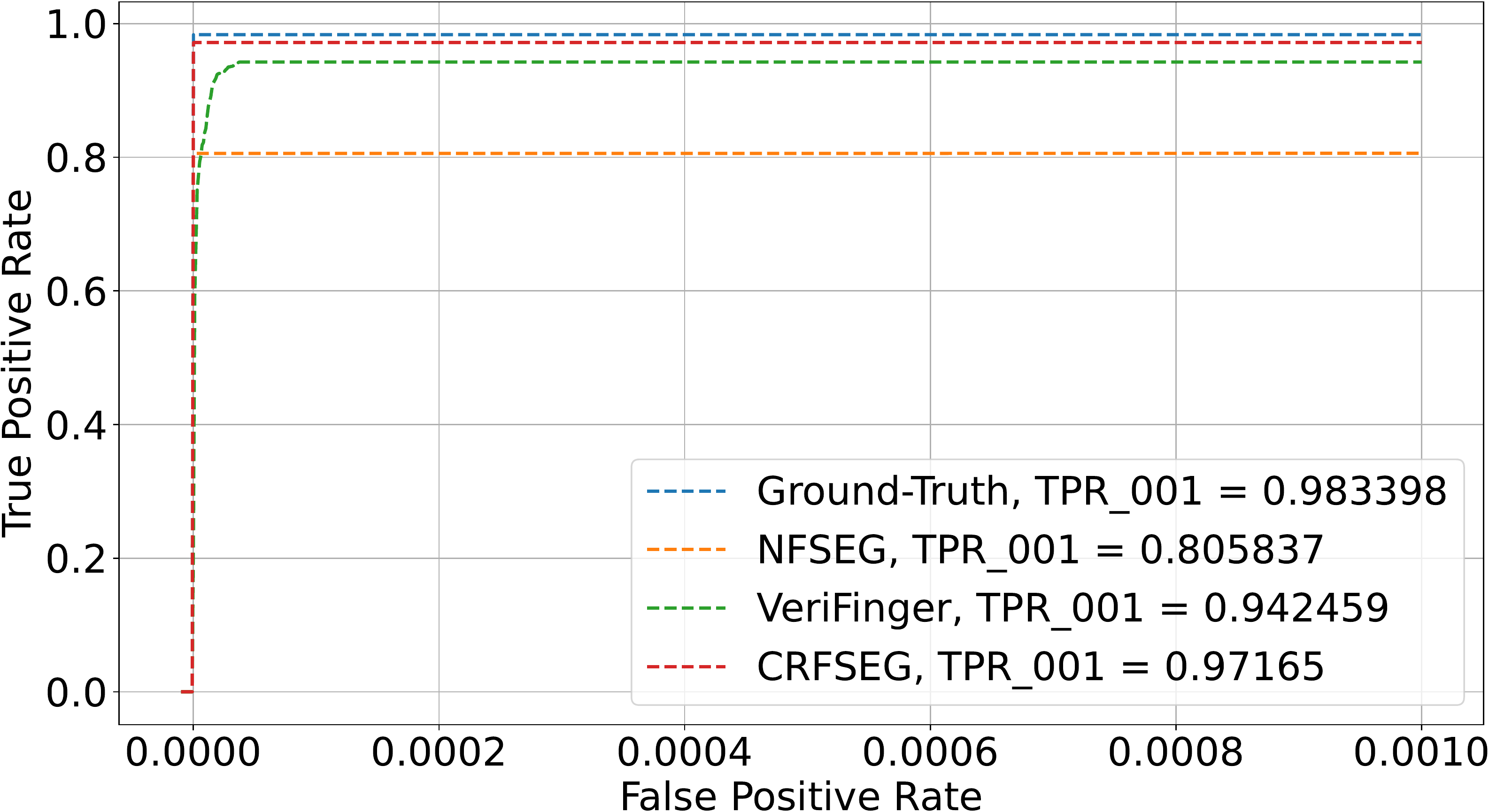}
\caption{The Receiver Operating Characteristics (ROC) for Ground Truth, NFSEG, VeriFinger, CRFSEG on both adult and children subjects in the \textit{Combined} dataset.}
\label{fig:matching_roc_entire}
\end{figure}

\autoref{table:MatchingAccuracy} lists the True Positive Rate (TPR) at False Positive Rate (FPR) of 0.001  for fingerprints segmented using bounding box information of ground-truth, NFSEG, VeriFinger, and CRFSEG. Ground truth achieves higher accuracy on both adult and children subjects. Among the segmentation algorithms, CRFSEG achieves better results compared to NFSEG and VeriFinger in terms of matching accuracy. 
\begin{table} [ht]
    \renewcommand{\arraystretch}{1}
\caption{True Positive Rate (TPR) at False Positive Rate (FPR) of 0.001 for fingerprints segmented using ground-truth, NFSEG, VeriFinger, and CFSEG on \textit{Combined} dataset. To evaluate model performance across different age groups, we calculate matching accuracies separately for adult and children subjects. The results indicate that CRFSEG performed close to the ground-truth level and outperformed VeriFinger and NFSEG in terms of fingerprint matching.}
\label{table:MatchingAccuracy} 
  \centering
  \begin{tabular}{ccccc}
    \hline 
{} & {} & {} & {Model} & {} \\
   \cline{3-5}
{} & {Ground-truth} & {NFSEG} & {VeriFinger} & {CRFSEG} \\
\cline{2-5}
 Dataset & Accuracy & Accuracy & Accuracy & Accuracy \\
\hline 
Children & \textbf{97.43}\% & {78.39}\% & {93.26}\% & {{96.47}}\% \\
\hline
Adult & \textbf{98.75}\% & {{82.31}}\% & {94.43}\% & {97.30}\% \\
\hline
Entire & \textbf{98.34}\% & {{80.58}}\% & {94.25}\% & {97.17}\% \\
\hline
\end{tabular}
\end{table}
\autoref{fig:matching_roc_children}, \autoref{fig:matching_roc_adult}, and \autoref{fig:matching_roc_entire} are three Receiver Operating Characteristics (ROC) curves that show the matching scores of different models along with ground-truth on children, adult, and entire slap dataset. 
The ROC curve is a graphical plot that represents the tradeoff
between the true positive rate (TPR) and the false positive rate
(FPR) at various threshold values. 
Our results indicate that for both adult and children subjects, the fingerprints segmented with CRFSEG provide higher accuracy among different segmentation algorithms across various threshold values. 

\subsubsection*{Label prediction accuracy on \textit{Challenging} dataset}
As we mentioned before, we created a dataset of slap images that NFSEG failed to segment, which we named the \textit{Challenging} dataset. We have finger information for each slap of that dataset but no bounding boxes around each fingerprint. Using finger information we calculated the label prediction accuracy of VeriFinger and CRFSEG on the \textit{Challenging} dataset.  \autoref{table:LabelPredictionOnDifficult} shows the results of label prediction accuracy, where CRFSEG outperformed VeriFinger by 9\% in terms of identifying fingers on the slap images. 

\begin{table} [ht]
    \renewcommand{\arraystretch}{1}
\caption{Label prediction accuracy of VeriFinger and CRFSEG on the \textit{Challenging} dataset. In this dataset, compared to VeriFinger, the CRFSEG model achieved better results by successfully segmenting all images with a label prediction accuracy of 91.83\%. This dataset is composed of slap images where NFSEG failed and thus did not have label accuracy.}
\label{table:LabelPredictionOnDifficult} 
  \centering
  \begin{tabular}{ccc}
    \hline 

Model & Dataset & Accuracy \\
\hline 
{VeriFinger} & {\textit{Challenging} Dataset} & {82.37\%} \\ \hline
{CRFSEG} & {\textit{Challenging} Dataset} & {91.83\%}\\\hline
\end{tabular}
\end{table}

\subsubsection*{Fingerprint matching results for \textit{Challenging} dataset}
As we do not have ground-truth bounding box information for the \textit{Challenging} dataset, we were not able to calculate the MAE or EAP of the different segmentation algorithms. However, we are able to calculate matching accuracy using the finger label information of the \textit{Challenging} dataset shown in \autoref{table:MatchingAccuracyOnDifficult}. 
\begin{table} [ht]
    \renewcommand{\arraystretch}{1}
\caption{TPR at FPR of 0.001 for VeriFinger, and CFSEG on the \textit{Challenging} dataset. The CRFSEG model demonstrated improved segmentation performance, as evidenced by its higher matching results compared to VeriFinger segmentation algorithm.}
\label{table:MatchingAccuracyOnDifficult} 
  \centering
  \begin{tabular}{ccc}
    \hline 

Model & Dataset & Matching Accuracy \\
\hline 
{VeriFinger} & {\textit{Challenging} Dataset} & {79.65\%} \\ \hline
{CRFSEG} & {\textit{Challenging} Dataset} & {89.93\%}\\\hline
\end{tabular}
\end{table}

\begin{figure}[!ht]
  \centering
  \includegraphics[width=0.9\linewidth]{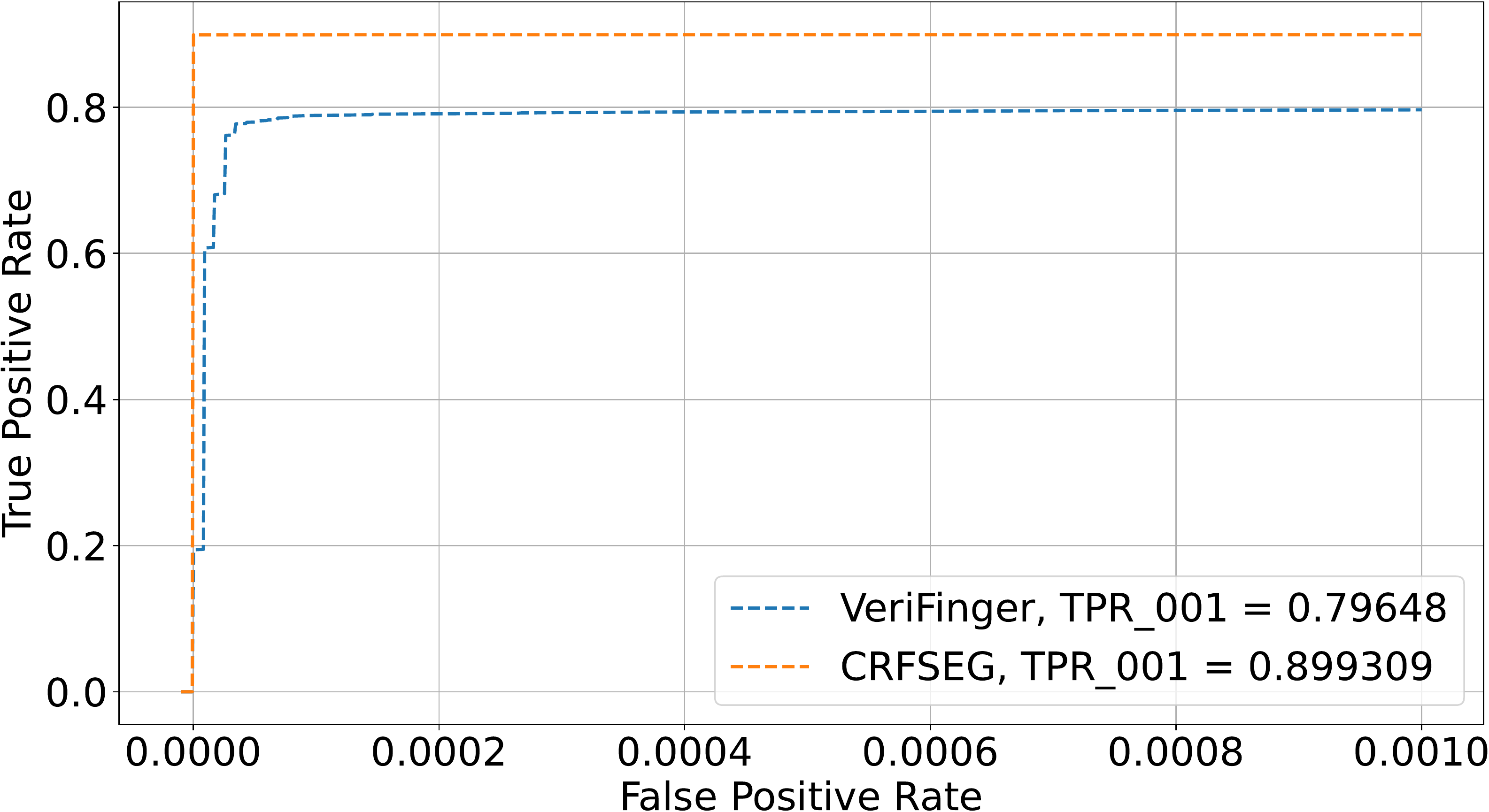}
\caption{Receiver Operating Characteristics (ROC) for VeriFinger, CRFSEG on the \textit{Challenging} dataset.}
\label{fig:matching_roc_entire_difficult}
\end{figure}

Out of 12370 fingerprints, VeriFinger segmented 10142 (82\%) fingerprints successfully, much less than the number of fingerprints segmented by CRFSEG which was 11338 (91\%). However, the VeriFinger matcher failed to process all the images segmented by the VeriFinger segmentation software and CRFSEG model. The reasons behind this failure are bad image quality, lack of fingerprint area, and fewer minutiae found on the segmented fingerprints. The number of failure images during matching was 2403 (21\%) for the VeriFinger segmentation software. On the other hand, the number of failure images during matching was 1485 (13\%) for the CRFSEG segmenter. This type of failure is considered a False Reject. 
The VeriFinger segmentation software had 2403 failure images during matching, which is 21\% of the total segmented fingerprint images of the \textit{Challenging} dataset, while the CRFSEG had 1485 failure images or 13\% of the total segmented fingerprint images.
\autoref{fig:matching_roc_entire_difficult} shows ROC for  VeriFinger, CRFSEG on the entire \textit{Challenging} image dataset.

\section*{Discussion}
This paper presents a highly accurate slap segmentation system constructed with a deep learning algorithm, which is capable of handling challenges such as rotation and other types of noise commonly found in slap images. The convolutional networks used in our system are trained using the end-to-end approach to achieve higher performance and are more generalized across different orientations of slap images. The advantage of our system is that once the entire segmentation system is built, it can be fine-tuned using additional datasets with fewer data points to facilitate the diversity of newer datasets. To the best of our knowledge, most commercial slap segmentation systems cannot easily be fine-tuned on different slap datasets.

In the slap dataset, we observed the presence of rotated slap images that posed a challenge for NFSEG to accurately calculate the rotational angle, resulting in less precise segmentation. This is a limiting factor to improving fingerprint-matching performance. To address this issue, we introduced augmented rotated images to our testing dataset and evaluated angle prediction performance.
Compared to commercially available slap segmentation methods like VeriFinger, our approach achieves  a lower error of $5.86^\circ$ in slap angle prediction, lower mean absolute error, and the highest fingerprint matching accuracy of 97.17\%. If we compare this matching accuracy to the human-annotated ground truth matching accuracy, our method achieves a very high score, indicating it performs at a human level. 

We observed that the label prediction accuracy of our CRFSEG is 0.21\% less than NFSEG on the \textit{Combined} dataset. However, NFSEG and VeriFinger systems require information about the hand, such as whether the slap image is of the left, right, or thumb, in order to accurately predict the labels of fingerprints within a slap image. On the contrary, our model predicts labels without any additional hand information. This property is particularly important when dealing with millions of fingerprints without additional information or fingerprints from a crime scene or any other source where visually it is not clear whether it is from the right hand or from the left hand. The ability of CRFSEG to label unknown fingers automatically, through its auto-labeling feature, results in a huge reduction in the time necessary for fingerprint matching.

An important point to consider in our novel segmentation model is that it was trained with fingerprints of different age groups. The main advantage of this is that it is robust toward different characteristics of fingerprints of different age groups. Indeed, our experimental results showed that our model is invariant to adult and children datasets. Moreover, we particularly focus on predicting arbitrarily oriented bounding boxes to increase the preciseness of segmented fingerprints. To achieve this objective, we designed an oriented region proposal network (ORPN) in our CNN architecture. The main advantage of this design is the model’s ability to learn fingerprint features with different orientations, resulting in better performance. The results of Mean Absolute Error (MAE) and Error in Angle Prediction (EAP) confirmed the notable superiority of our proposed approach in terms of bounding box predictions. 

Our deep learning-based segmentation system has some limitations, including its dependency on large amounts of labeled data for training, as well as the need for significant computational resources for training and inference, which may hinder its scalability and accessibility.

During testing on the challenging dataset, we discovered that the training dataset used to train CRFSEG did not include any annotated slap images with missing fingers, which were present in the challenging dataset. As a result of being trained on a dataset without missing fingerprints, CRFSEG occasionally mislabels fingerprints in slap images, even though it can accurately detect the fingerprints of such images.
This problem can be addressed by adding images with missing fingers to the training dataset. In summary, CRFSEG may struggle to segment new and diverse data beyond the scope of the training dataset.  

\section*{Conclusions}
We designed an oriented bounding box-based deep learning method to segment overly rotated slap fingerprint images of different age groups. 
Unlike conventional image detection and segmentation algorithms, this algorithm uses five parameters to determine a rotated or axis-aligned object, where four parameters are used to determine the position of a bounding box around the object, and the last one is used for the angle of that bounding box. 

We used a modified version of Faster R-CNN to achieve this goal. The two-stage object detection architecture depicted in \autoref{fig:rotatedFasterRCNNArchi} utilizes a deep learning feature extractor model, pre-trained on the Imagenet dataset, to extract semantic-rich feature maps from input slap images at different scales. These feature maps are then processed by the O-RPN, a modified regional proposal network specially designed for detecting both axis-aligned and rotated objects, to precisely segment objects in an input image. Another branch of the detection architecture uses the same feature maps but outputs the label of objects.

In the training stage, the dataset was initially augmented to increase its size and to incorporate common scenarios found in a real-world slap dataset, such as overly rotated slap images.  Then, the Faster R-CNN based search approach was used to find the positive ROIs by calculating the IOU between detected bounding boxes and ground-truth bounding boxes. Those positive ROIs were used to learn the fingerprint features in a slap image. 
A scale-up search procedure is used during training to find positive ROIs and to ensure that all potential fingerprint regions are found and processed correctly.
In the inference stage, all the test slap images are processed by the feature extraction network, and hundreds of ROIs are proposed by the O-RPN. After that, the cross-entropy is performed to generate class labels, and regression is performed to locate the precise bounding box around fingerprints. A threshold ($\ge 0.7$) of the class score is used to filter the proposal regions and determine whether it is a fingerprint. Finally, post-processing, such as non-maximum suppression, is performed to obtain the final segmentation. 

We segmented all the slap images using ground-truth information, NFSEG, VeriFinger, and CRFSEG, resulting in four sets of segmented fingerprint images.
The success of various fingerprint segmentation systems was evaluated by comparing the matching scores of the segmented images produced by each system. The CRFSEG model showed a 97.17\% match accuracy on the \textit{Combined} dataset, which was higher compared to the NFSEG and Verifinger systems, which had 80.58\% and 94.25\% match accuracy respectively. We also calculated the MEA, EAP, and label prediction accuracy to evaluate all the slap segmentation models.
The experimental result showed that our CRFSEG model segmented fingerprints robustly across over-rotated slap images for both children and adult subjects. We outperformed state-of-the-art NFSEG and VeriFinger in terms of MAE, EAP, and matching performance. A pre-trained CRFSEG model and corresponding training codes are publicly accessible (\url{https://github.com/sarwarmurshed/CRFSEG}). 

\printbibliography

\section*{Acknowledgements}
This material is based upon work supported by the Center for Identification Technology Research and the National Science Foundation (NSF) under Grant No.$1650503$. We would like to thank Winnie Liu and Heidi Walko for their contribution in preparing the dataset. We also extend their gratitude to Precise Biometrics and the Potsdam Elementary and Middle School administration, staff, students, and the parents of the participants for supporting our research and the greater goal of scientific contribution to society. The authors would also like to thank the \textit{Chameleon} cloud for providing the computational infrastructure required for this work \cite{keahey_lessons_2020}.

\end{document}